\documentclass[preprint,12pt]{elsarticle}

\usepackage{amssymb}
\usepackage{algorithm}%
\usepackage{algorithmicx}%
\usepackage{algpseudocode}%
\usepackage{listings}%
\usepackage{graphicx}
\usepackage{amsmath}
\usepackage{booktabs}
\usepackage{array} 
\usepackage{subcaption}
\usepackage{multirow}  
\usepackage{graphicx}
\usepackage{hyperref}
\usepackage[toc]{appendix}
\UseRawInputEncoding

\journal{Engineering Applications of Artificial Intelligence}

\begin{document}

\begin{frontmatter}



\title{Mamba Adaptive Anomaly Transformer with association discrepancy for time series}


\author[a,c]{Abdellah Zakaria Sellam}
\author[a]{Ilyes Benaissa}
\author[b]{Abdelmalik Taleb-Ahmed}
\author[a,c]{Luigi Patrono}
\author[a,c]{Cosimo Distante}

\affiliation[a]{organization={Department of Innovation Engineering; University of Salento},
            addressline={via Monteroni}, 
            city={Lecce},
            postcode={73100}, 
            country={Italy}}

\affiliation[b]{organization={Institute d'Electronique de Microelectronique et de Nanotechnologie (IEMN); UMR 8520; Universite Polytechnique Hauts de France; Universite de Lille; CNRS},
            addressline={}, 
            city={Valenciennes},
            postcode={59313}, 
            country={France}}

\affiliation[c]{organization={Institute of Applied Sciences and Intelligent Systems - CNR},
            addressline={via Monteroni}, 
            city={Lecce},
            postcode={73100}, 
            country={Italy}}

\begin{abstract}
Anomaly detection in time series poses a critical challenge in industrial monitoring, environmental sensing, and infrastructure reliability, where accurately distinguishing anomalies from complex temporal patterns remains an open problem. While existing methods, such as the Anomaly Transformer leveraging multi-layer association discrepancy between prior and series distributions and DCdetector employing dual-attention contrastive learning, have advanced the field, critical limitations persist. These include sensitivity to short-term context windows, computational inefficiency, and degraded performance under noisy and non-stationary real-world conditions. To address these challenges, we present MAAT (Mamba Adaptive Anomaly Transformer), an enhanced architecture that refines association discrepancy modeling and reconstruction quality for more robust anomaly detection.
Our work introduces two key contributions to the existing Anomaly transformer architecture: Sparse Attention, which computes association discrepancy more efficiently by selectively focusing on the most relevant time steps. This reduces computational redundancy while effectively capturing long-range dependencies critical for discerning subtle anomalies. A Mamba-Selective State Space Model (Mamba-SSM) is also integrated into the reconstruction module. A skip connection bridges the original reconstruction and the Mamba-SSM output, while a Gated Attention mechanism adaptively fuses features from both pathways. This design balances fidelity and contextual enhancement dynamically, improving anomaly localization and overall detection performance.
Extensive experiments on benchmark datasets demonstrate that MAAT significantly outperforms prior methods, achieving superior anomaly distinguishability and generalization across diverse time series applications. By addressing the limitations of existing approaches, MAAT sets a new standard for unsupervised time series anomaly detection in real-world scenarios.
You can find the link to the GitHub account in the link provided below.
\url{https://github.com/ilyesbenaissa/MAAT}
\end{abstract}

\begin{graphicalabstract}
\includegraphics{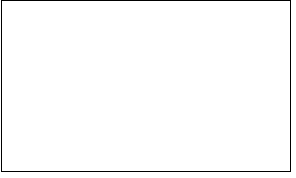}
\end{graphicalabstract}

\begin{highlights}


    \item The Mamba Adaptive Anomaly Transformer (MAAT) features a \textbf{Sparse Attention} mechanism. It incorporates the \textbf{Mamba-Selective State Space Model (Mamba-SSM)}, enabling it to effectively capture both \textbf{short-term and long-term} temporal dependencies in time series data.
    \item Utilizing a \textbf{block-wise sparse attention} approach to minimize computational costs while enhancing anomaly detection performance.
    \item Applying a \textbf{gated attention} to merge sparse attention with the outputs of the Mamba-SSM, thereby improving both anomaly localization and detection.
    \item Leverages the \textbf{State Space Model (SSM)} through selective use of Mamba, facilitating more \textbf{effective management of complex, noisy data}.
    \item MAAT \textbf{outperforms} existing models such as the \textbf{Anomaly Transformer} and \textbf{DCdetector}, achieving significantly higher \textbf{F1 scores, accuracy, and recall} across various benchmark datasets.
\end{highlights}


\end{frontmatter}


\section{Introduction}
\label{sec:introduction}
Anomaly detection in time series data is critical in finance, healthcare, industrial monitoring, and cybersecurity to prevent failures, detect fraud, and ensure efficiency. Traditionally, methods like ARIMA \cite{box1970time} and Gaussian Processes \cite{rasmussen2006gaussian} have been used to identify abnormal events in sequential data. Which rely on the assumption that anomalies can be identified as deviations from predicted values. However, these methods often struggle with real-world time series data of a complex, high-dimensional nature, especially when the underlying patterns are non-linear or when the anomalies are subtle\cite{box2015time}.
Machine learning introduced advanced models like Support Vector Machines (SVMs)\cite{scholkopf2000support}, Random Forests\cite{rabiner1986introduction}, and Hidden Markov Models (HMMs)\cite{hochreiter1997long} were among the first machine learning methods applied for time series anomaly detection, improving on traditional methods by learning from data. However, they still required extensive feature engineering and often missed complex temporal dependencies.
Deep learning models such as Recurrent Neural Networks (RNNs)\cite{radford2018network}, and Long Short-Term Memory (LSTM)\cite{malhotra2016lstm} revolutionized time series anomaly detection by capturing long-term dependencies and complex temporal patterns, ideal for identifying subtle deviations \cite{hochreiter1997long}. Additionally, the use of autoencoders and their variants, such as Variational Autoencoders (VAEs)\cite{xu2018unsupervised} and Adversarial Autoencoders\cite{somepalli2021unsupervised}, has further advanced the field by enabling unsupervised anomaly detection through the reconstruction of standard data patterns \cite{kingma2013auto}.
Reconstruction-based methods, like autoencoders \cite{malhotra2016lstm}, model standard data patterns and detect anomalies through reconstruction errors, leveraging their ability to learn compressed representations \cite{hinton2006reducing}. Similarly, recurrent neural networks (RNNs) and long short-term memory (LSTM) networks have been employed to capture temporal dependencies in sequential data during reconstruction tasks \cite{malhotra2016lstm}. However, reliance on reconstruction loss can cause false positives for anomalies resembling standard data. Transformers\cite{vaswani2017attention}, with self-attention mechanisms, address this by modeling complex dependencies in time series. Enhancements like gated attention \cite{gating_attention}, for instance, integrate gating structures to dynamically control the flow of information dynamically, enabling the model to focus on relevant temporal features and improve performance in time series forecasting tasks. 
Sparse attention\cite{sparse_attention} mechanisms have also been introduced to improve the efficiency of Transformers when handling long sequences. By limiting the attention computation to a subset of the input sequence, sparse Attention reduces computational complexity while maintaining the ability to capture essential dependencies. This approach is particularly beneficial for long-sequence time-series forecasting, where traditional attention mechanisms may become computationally prohibitive. 
Recent advancements include the Anomaly Transformer \cite{xu2022anomaly}, which uses association discrepancy learning to differentiate normal from abnormal points without relying solely on reconstruction loss. It employs Gaussian kernels and attention mechanisms to model prior- and series-association discrepancies, improving accuracy Accuracystness while addressing scalability in high-dimensional datasets. Despite achieving state-of-the-art performance through a minimax strategy, it faces limitations: self-attention struggles with long-range dependencies in small windows, and noise or non-stationary patterns can increase false positives during normal fluctuations. While more efficient than traditional methods, these challenges remain.
Furthermore, DCdetector \cite{yang2023dcdetector} simplifies anomaly detection with a dual-attention contrastive structure, eliminating complex components like Gaussian kernels or reconstruction losses. It uses contrastive learning to separate anomalies from expected points, capturing global and local dependencies through a dual-channel architecture. Its channel-independent patching reduces parameter complexity and overfitting risks, while a contrastive loss function based on Kullback-Leibler divergence enhances representation consistency. However, DCdetector faces challenges, including sensitivity to contrastive sample quality, high computational overhead from pairwise comparisons, and dual-branch parallelism.
Building upon these foundations, we propose MAAT (Mamba Adaptive Anomaly Transformer with association discrepancy for time series) for anomaly detection in time series. Our approach introduces a novel block architecture that incorporates skip connections and gating mechanisms in
association with the mamba block in the skip connection to improve the reconstruction capability of the anomaly transformer architecture. By combining these architectural enhancements with sparse attention mechanisms and anomaly Transformer association modeling principles, MAAT performs better in detecting anomalies across diverse datasets.
The Key contributions of this work are summarized as follows:
\begin{itemize}
    \item integrated sparse attention mechanisms to replace the standard attention mechanism in the original Anomaly Transformer, allowing scalable and efficient processing of long time series data. By enabling dynamic control over block size, MAAT achieves an optimal balance between computational efficiency and precision in anomaly detection.
    
    \item Introduced the MAMBA block to the Anomaly Transformer architecture to optimize the reconstruction of signals.
    \item Incorporated the MAMBA block with skip connections and gating Attention to enhance the reconstruction output and reduce the loss of the Anomaly Transformer architecture.
\end{itemize}
The subsequent sections of this document are structured as follows:
Section 2 presents a comprehensive overview of relevant works, including literature and methodologies, related to time series anomaly detection. Section 3 delves into our proposed MAAT architecture, detailing
its unique features and potential benefits. In Section 4, we conduct
Experiments, perform data analysis, and provide a thorough evaluation of our model. Lastly, in Section 5, we draw insightful conclusions based on the following:
In our experimentation, we compare our approach with prior methodologies and articulate the implications of our findings. This structure
ensures a coherent and comprehensive understanding of our innovative
methodology for unsupervised anomaly detection in time series data. 
\section{Related works}
Time series anomaly detection has evolved from traditional statistical models like ARIMA and Gaussian Processes\cite{box2015time,rasmussen2006gaussian}. Conventional methods like ARIMA and GP struggled with non-linear and high-dimensional data. Machine learning introduced more flexible approaches, such as  (SVMs), particularly One-Class SVM (OC-SVM), which excel in novelty detection by defining a boundary around standard data and identifying outliers as anomalies \cite{scholkopf2000support}. Despite this improvement, SVMs could not capture temporal dependencies inherent in sequential data.
Random Forests and  Isolation Forests handled high-dimensional data better than SVMs, while Isolation Forests identified anomalies by isolating points through recursive partitioning \cite{liu2008isolation}. However, they focused on anomalies and were not optimized for capturing time dependencies in sequential data. RNNs and  LSTMs became popular due to their ability to capture long-term dependencies in time series data. These models excelled in detecting anomalies in healthcare (e.g., ECG data) and finance \cite{distante2022hf}. However, RNNs and LSTMs face limitations in scalability and can struggle with vanishing gradient problems when processing long sequences.
Variational Autoencoders (VAEs) became central to unsupervised anomaly detection by reconstructing standard data patterns and flagging high-reconstruction errors as anomalies \cite{kingma2013auto}. Adversarial Autoencoders (AAEs) and Generative Adversarial Networks (GANs) enhanced robustness through adversarial training. GANs use a generator-discriminator framework to detect anomalies based on realistic data generation. However, both GANs and AEs require large datasets and can face instability, limiting their application in some domains \cite{schlegl2019f,li2019mad}.
Transformer-based models have recently demonstrated significant potential in time series anomaly detection by leveraging self-attention mechanisms to capture long-range dependencies \cite{vaswani2017attention}. Initially developed for NLP, these models eliminate the need for recurrent structures like RNNs or LSTMs, effectively modeling local and global patterns in time series data to detect anomalies.
Transformer-based models have revolutionized time series anomaly detection, with innovations like the Anomaly Transformer introducing Association Discrepancy to compare expected and observed associations in data. Using a minimax strategy enhances the distinguishability of anomalies, outperforming prior methods across datasets \cite{xu2022anomaly}. Similarly, AnomalyBERT employs self-supervised learning and data degradation to simulate anomalies, improving generalization without labeled data \cite{jeong2023anomalybert}. The Denoising Diffusion Mask Transformer (DDMT) integrates denoising diffusion with masking mechanisms, excelling in noisy multivariate settings \cite{yang2023ddmt}.
Multivariate anomaly detection has also advanced with models like Informer \cite{zhou2021informer} and multi-task Transformers, which leverage attention mechanisms to model interdependencies among variables \cite{zhang2021multitask}. Hybrid models combining statistical methods with deep learning enhance interpretability while maintaining flexibility. These developments address the limitations of classical met techniques as ARIMA, which struggled with high-dimensional, non-stationary data.
As the field progressed, deep learning frameworks began to emerge. In 2021, the \textbf{Anomaly Transformer}~\cite{xu2022anomaly} introduced a novel anomaly-attention mechanism that quantifies the association discrepancy between normal and abnormal points. While this approach effectively leverages self-attention to capture both local and global temporal dependencies in an unsupervised manner, its quadratic complexity poses challenges for scalability, and its underlying assumptions may not hold across all types of anomalies.
Building on these advances, the \textbf{DC Detector}~\cite{yang2023dcdetector} employs a dual attention contrastive representation learning framework. Integrating local and global Attention with a contrastive loss overcomes some of the pitfalls of reconstruction-based methods. Nonetheless, the additional architectural complexity and the need for extensive hyperparameter tuning remain nontrivial hurdles.
Concurrently, selective state space models have gained traction for their efficiency. Models like \textbf{MAMBA}~\cite{gu2023mamba} use a selective scanning mechanism to model long-range dependencies linearly, making them attractive for real-time and large-scale applications. However, the trade-off is a potential flexibility reduction when modeling highly non-linear interactions.
Despite these significant advances, several challenges persist. One major issue is \textbf{concept drift}, the phenomenon where the underlying distribution of data evolves. This drift renders pre-trained models ineffective and highlights the need for adaptive learning techniques that can update dynamically as new data becomes available. In addition, scalability remains a critical challenge, especially when dealing with high-frequency time series data or real-time anomaly detection in large-scale systems. Recent efforts in developing memory-efficient Transformers and distributed learning approaches~\cite{gupta2021memory,lai2021revisiting} have begun to address these scalability issues, marking promising steps toward more adaptive and robust anomaly detection systems.
Overall, the evolution of unsupervised time series anomaly detection reflects a broader trajectory across complex domains. The field is steadily progressing toward more nuanced, robust, and adaptable solutions, from early statistical methods to modern deep learning architectures that harness attention mechanisms, contrastive learning, and efficient state space representations. These advancements enhance our ability to capture intricate temporal dependencies and subtle deviations without relying on labeled data and pave the way for deploying these models in real-world, dynamic environments.
\section{Methodology}
\begin{figure*}[h]
\centering
\includegraphics[width=1.1\textwidth]{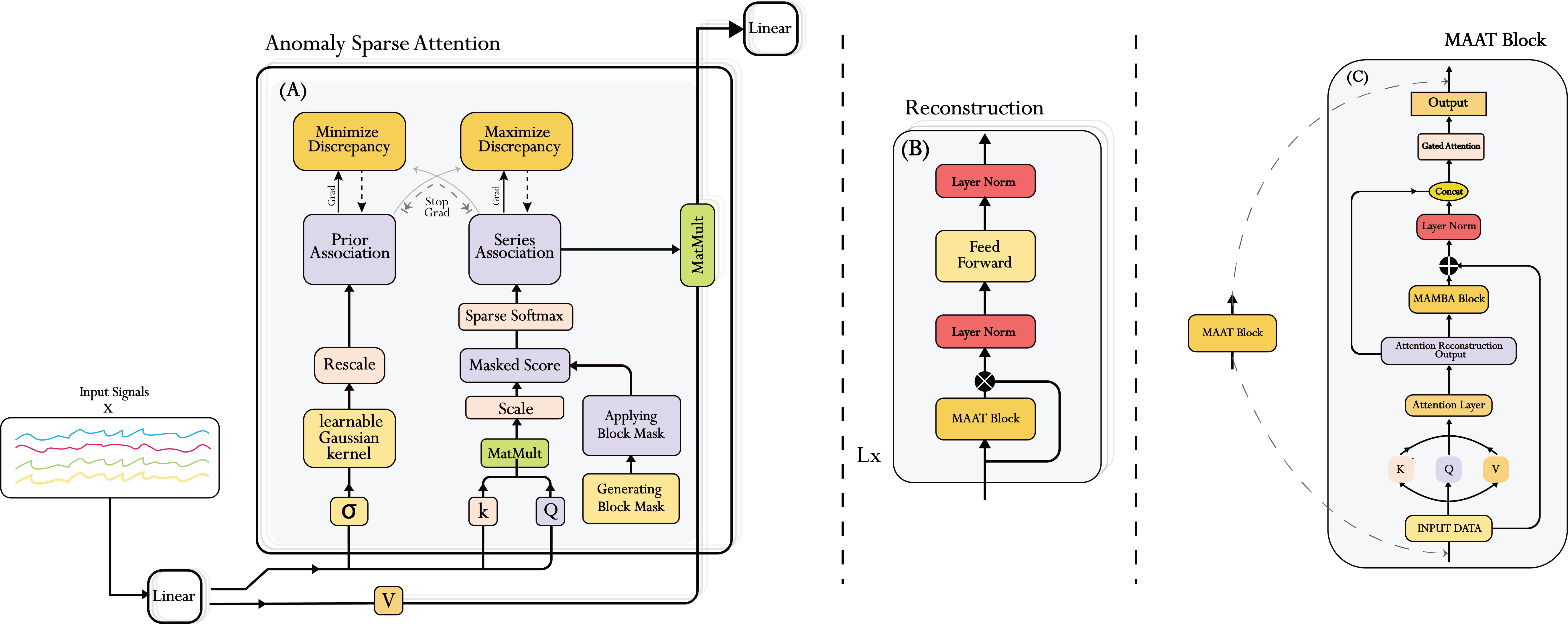}
\caption{The figure illustrates the MAAT framework for time series anomaly detection. Block A (Anomaly Sparse Attention Module) computes prior and series associations using sparse Attention and a learnable Gaussian kernel to model temporal dependencies. Block B (Reconstruction Module) refines the feature representations using layer normalization, feedforward processing, and MAAT blocks to reconstruct input signals effectively. Block C represents the MAAT Block that integrates the Mamba state-space model to capture long-range dependencies, followed by a gated attention mechanism that adaptively fuses the reconstructed output.}
\label{fig:maat}
\end{figure*}

Consider a system that records a sequence of \( d \)-dimensional measurements at uniform time intervals. The time series data is represented by the set \(\{x_1, x_2, \dots, x_N\}\), where each \( x_t \in \mathbb{R}^d \). The goal is to determine whether an observation \( x_t \) is anomalous in an unsupervised manner.
Effective unsupervised time series anomaly detection relies on learning informative representations and establishing a clear discriminative criterion. The original Anomaly Transformer framework distinguishes standard patterns from anomalies by learning an association discrepancy via anomaly attention and a minimax optimization strategy. Building on this, the Mamba Adaptive Anomaly Transformer (MAAT) introduces Anomaly Sparse Attention, Mamba Blocks, and Gated Skip Connections, enhancing short-range dependency modeling and long-range temporal learning while reducing computational cost.
As illustrated in Figure~\ref{fig:maat}(A), Anomaly Sparse Attention replaces dense self-attention with a two-branch mechanism. The Prior-Association Branch encodes expected dependencies using a learnable Gaussian kernel, while the Series-Association Branch applies block-wise sparse Attention to capture time series patterns adaptively. These associations guide discrepancy learning, enhancing anomaly detection performance.
The Reconstruction Block, shown in Figure~\ref{fig:maat}(B), refines extracted features by stacking MAAT Blocks with LayerNorm and Feedforward Networks (FFN). This layered design supports hierarchical feature learning, making the model more robust to diverse time series behaviors.
Finally, the MAAT Block, depicted in Figure~\ref{fig:maat}(C), integrates Mamba Blocks within a Gated Attention mechanism, selectively amplifying meaningful signals while filtering out noise, and the skip connection preserves critical information while adapting the model focus.
MAAT provides a scalable and expressive framework for detecting complex anomalies across various time series datasets by combining sparse Attention, state-space modeling, and gated learning mechanisms with the association discrepancy.

\subsection{Background knowledge}
\subsubsection{Association Discrepancy}
The Anomaly Transformer framework employs a minimax optimization strategy that alternates between two phases to balance the learning of two complementary feature sets, \(P\) and \(S\). This alternating approach prevents collapse in the representation of \(P\) while ensuring that \(S\) captures non-local dependencies.
\paragraph{Minimize Phase:} 
As introduced in the Anomaly Transformer framework, this minimax strategy ensures that both feature sets capture complementary aspects of the input data, enhancing the overall quality of the learned representations. The hyperparameter \(\lambda\) is critical in balancing the network's ability to reconstruct the input with the need for diverse and informative feature learning.
\subsubsection{Association-based Anomaly Criterion}  
The Anomaly Transformer framework integrates temporal pattern analysis with reconstruction fidelity through a dual-mechanism scoring system. The final anomaly score for an input \( X \in \mathbb{R}^{N \times d} \) is computed as:  
\begin{equation}
\text{AnomalyScore}(X) = \text{Softmax}\Bigl(-\text{AssDis}(P, S; X)\Bigr) \odot \| X_{i,:} - \hat{X}_{i,:} \|_2^2, \quad i = 1, \ldots, N,
\label{eq:anomaly_score}
\end{equation}  
Where:  
\begin{itemize}
    \item \(\text{AssDis}(P, S; X)\) measures the association discrepancy between prior-association (\(P\)) and series-association (\(S\)),
    \item \(\text{Softmax}(\cdot)\) normalizes the negative association discrepancy across time steps,
    \item \(\odot\) represents element-wise multiplication,
    \item \(\| X_{i,:} - \hat{X}_{i,:} \|_2^2\) quantifies the reconstruction error at each time step \(i\).
\end{itemize}
This formulation combines temporal patterns and reconstruction accuracy for anomaly detection. While improved reconstructions reduce discrepancy, anomalies with significant deviations in reconstruction or temporal patterns still yield high scores. Probabilistic weighting highlights time steps with poor reconstruction and abnormal dependencies, enhancing robustness.
\subsection{Sparse Attention for Efficient Sequence Attentiong}
Sparse Attention restricts token interactions by applying a sparsity pattern \( S \) that selects only a subset of pairs for attention computation \cite{child2019generating, zaheer2020big, kitaev2020reformer}.
Sparse attention introduces a sparsity mask \( S \) to limit interactions:
\begin{equation}
\text{SparseAttention}(Q, K, V) = \text{softmax}\left(\frac{(QK^T) \odot S}{\sqrt{d_k}}\right)V,
\label{eq:sparse_attention}
\end{equation}
Where:
\begin{itemize}
    \item \( Q, K, V \in \mathbb{R}^{N \times d} \) are the query, key, and value matrices,
    \item \( d_k \) denotes the dimensionality of the key vectors,
    \item The softmax function normalizes the attention scores.
\end{itemize}

where \( \odot \) represents element-wise multiplication. This ensures that only selected token pairs contribute to the attention mechanism.

Approaches for defining the sparsity pattern \( S \) include:
\begin{itemize}
    \item \textbf{Fixed Patterns:} Predefined structures, such as strided or block-wise masks \cite{child2019generating}.
    \item \textbf{Learned Masks:} Patterns that are dynamically optimized during training \cite{zaheer2020big}.
    \item \textbf{Memory-Efficient Methods:} Techniques like locality-sensitive hashing (LSH) that enable content-based token selection \cite{kitaev2020reformer}.
\end{itemize}




The Gated Attention Mechanism enhances standard attention by using a learnable gating process to dynamically adjust the importance of input features. It evaluates the relevance of features to optimize attention weight distribution, represented mathematically as:
\[
\text{GatedAttention}(x) = \sigma\bigl(G(x)\bigr) \odot A(x)
\]
Key components include \(x\)input data, \(G(x)\)gating vector from a neural network,\(\sigma\): sigmoid function mapping to \([0,1]\),\(A(x)\) standard attention weights,
\(\odot\)element-wise multiplication.
This mechanism allows the model to focus on task-critical features while suppressing noise, improving robustness in contexts like anomaly detection.
\subsection{Mamba State Space Model}
State Space Models (SSMs) are widely used in time series modeling because they capture latent dynamics efficiently. The \textbf{Mamba State Space Model (Mamba SSM)} is a recent advancement in structured state-space models (S4) designed for long-range sequence modeling while maintaining computational efficiency \cite{gu2022efficiently, tay2023mamba}.
The following equations define Mamba SSM:

\subsection{Mamba Adaptive Anomaly Transformer Architecture}

Figure~\ref{fig:maat} illustrates the overall structure of the Mamba Adaptive Anomaly Transformer (MAAT), an architecture inspired by the Anomaly Transformer. MAAT integrates a sparse attention mechanism, a selective state-space model known as the Mamba block, and an adaptive reconstruction framework. The sparse attention mechanism processes the input time series data, capturing local dependencies efficiently. The Mamba block is designed to capture long-range dependencies, enhancing the model's ability to understand complex temporal patterns. An adaptive gating mechanism then fuses the outpuAttentionhe sparse attention and Mamba block, enabling the model to reconstruct the input signal effectively. This combination allows MAAT to detect anomalies by comparing the reconstructed signal with the original input, identifying deviations that may indicate anomalous behavior.

\subsubsection{Anomaly sparse attention}
Our anomaly sparse attention module, shown in Figure~\ref{fig:maat}(A), refines the conventional anomaly attention used in the Anomaly Transformer, which relies on self‐attention  and sparse Attention. While the prior-association remains unchanged, modeled using a learnable Gaussian kernelthe series-association is now computed through sparse self-attention, capturing observed dependencies more efficiently.

In this module, the prior association continues to be represented by a learnable Gaussian kernel, while the series association is computed using a sparse softmax operation over local attention windows.

For each query vector \( Q_i \), normalization is performed only over keys in the local window defined by  
\begin{equation}
\Omega_i = \{ j \mid |j-i| \leq \text{block\_size}/2 \},
\label{eq:omega_i}
\end{equation}  

so that the series association is computed as  
\begin{equation}
S^l_{i,j} = \frac{\exp\left(\frac{Q_i K_j^T}{\sqrt{d_{\text{model}}}}\right)}{\sum_{k \in \Omega_i} \exp\left(\frac{Q_i K_k^T}{\sqrt{d_{\text{model}}}}\right)}, \quad \forall\, j \in \Omega_i,
\label{eq:s_l_ij}
\end{equation}  

with  
\begin{equation}
S^l_{i,j} = 0 \quad \text{for } j \notin \Omega_i.
\label{eq:s_l_ij_zero}
\end{equation}  

This sparse softmax formulation ensures that only locally relevant keys are considered during normalization, reducing computational redundancy while preserving essential dependencies.

\subsubsection{MAAT Adaptive Block}
The adaptive block shown in Figure \ref{alg:maat}(C) integrates a state-space model (Mamba block) with a skip connection that retains the original input. The Mamba block is specifically designed to capture long-range dependencies, while the skip connection ensures the preservation of fine details. We achieve a robust intermediate representation by combining these elements and applying layer normalization.
This Adaptive Block merges long-range dependencies and local features through a state-space (Mamba) skip path and an adaptive gating mechanism. After performing sparse attention processing, we proceed with the following steps:
The Mamba block generates a transformed representation \( x_{Mamba{mamba}} \), which is then combined with the reconstructed input from sparse attention \( x_{\text{orig}} \) via a residual connection. This output is normalized as follows:
\begin{equation}
x_{\text{skip}} = \text{LayerNorm}\bigl(x_{\text{mamba}} + x_{\text{orig}}\bigr),
\label{eq:x_skip}
\end{equation}

\paragraph{Adaptive Gating and Reconstruction:}  
In the context of adaptive gating and reconstruction, a gating factor \( g \) is computed from the concatenation of the output from the central processing path \( x \) and the corresponding skip connection \( x_{\text{skip}} \). Formally, the gating factor can be expressed as follows:
\begin{equation}
g = \sigma\Bigl(\text{Linear}\bigl([x; \, x_{\text{skip}}]\bigr)\Bigr) = \sigma\Bigl(W [x; \, x_{\text{skip}}] + b\Bigr),
\label{eq:g}
\end{equation}

Where \( [x; \, x_{\text{skip}}] \) denotes the concatenation of the feature maps along the channel dimension, \( W \) and \( b \) represent the learnable parameters of the linear transformation (precisely, the weight matrix and bias term), and \( \sigma(\cdot) \) is the sigmoid activation function that maps the output to the range \((0,1)\).

The gating factor \( g \) is critical in modulating the relative importance of the skip and main paths. Specifically, for each element:
\begin{itemize}
    \item When \( g \) approaches 1, the output is predominantly influenced by the skip connection \( x_{\text{skip}} \), which encapsulates long-range contextual information along with a residual signal from the original input.
    \item Conversely, when \( g \) approaches 0, the reconstruction is primarily derived from the main path output \( x \), which reflects the locally processed features.
\end{itemize}

The final output, denoted as \( X^{\text{adapt}} \), is computed as an element-wise weighted combination of these two representations:
\begin{equation}
X^{\text{adapt}} = g \odot x_{\text{skip}} + (1-g) \odot x,
\label{eq:x_adapt}
\end{equation}

where \( \odot \) signifies element-wise multiplication. This formulation enables the network to dynamically calibrate the contributions from the skip connection and the main processing branch, thereby facilitating a harmonious integration of local details and global context. The process of adaptive fusion is paramount in ensuring that the reconstructed output \( X^{\text{adapt}} \) accurately embodies normal patterns. In contrast, anomalies disrupt local consistency and global structure, increasing reconstruction errors. This enhanced reconstruction capability, in turn, fosters improved performance in anomaly detection.

Consequently, \( X^{\text{adapt}} \) represents a sophisticated integrated reconstruction, merging information from both the skip path and the central processing branch, ultimately enhancing the overall representational capacity of the model.

The anomaly score now leverages the adaptively fused reconstruction \( X^{\text{adapt}} \) by balancing the association discrepancy from Eq.~\eqref{eq:ass_dis_updated} and the reconstruction error. Specifically, the anomaly score for each time step \( i \) is defined as
\begin{equation}
    \label{eq:ass_dis_updated}
    \text{AnomalyScore}(X) = \text{Softmax} \Bigl( -\text{AssDis}(P, S; X) \Bigr) \odot \Bigl\| X_{i,:} - X^{\text{adapt}}_{i,:} \Bigr\|_2^2, \quad i=1, \dots, N.
\end{equation}

Here, \( \odot \) represents element-wise multiplication, and:
\begin{itemize}
    \item \( \text{AnomalyScore}(X) \in \mathbb{R}^{N \times 1} \) provides the point-wise anomaly criterion.
    \item \( \text{AssDis}(P, S; X) \) is the association discrepancy, as defined in Eq.~\eqref{eq:ass_dis_updated}.
    \item \( \| X_{i,:} - X^{\text{adapt}}_{i,:} \|_2^2 \) is the reconstruction error between the input \( X \) and its adaptively reconstructed counterpart \( X^{\text{adapt}} \).
\end{itemize}
This updated formulation clearly distinguishes the adaptively reconstructed output \( X^{\text{adapt}} \), which utilizes the gating for reconstruction from the input \( X \), emphasizing the role of adaptive fusion in our anomaly detection criterion.

\section{Experiments}
\subsection{Benchmark Datasets}
In this study, we evaluate the performance of our model using eight representative benchmarks derived from five real-world applications. The first dataset, MSL, is the Mars Science Laboratory dataset collected by NASA, which reflects the condition of sensors and actuator data from the Mars rover \cite{MSL_dataset}. Similarly, the SMAP dataset, provided by NASA, presents soil samples and telemetry information from the Mars rover; notably, SMAP contains more point anomalies than MSL \cite{SMAP_dataset}. 
The PSM dataset, a public resource from eBay Server Machines, includes 25 dimensions and is widely used for research in anomaly detection \cite{PSM_dataset}. Additionally, the SMD dataset consists of a five-week-long record of resource utilization traces collected from an internet company compute cluster, monitoring 28 machines \cite{SMD_dataset}. 
Another critical benchmark is the SWaT dataset, which comprises 51-dimensional sensor data from a secure water treatment system that operates continuously \cite{SWaT_dataset}. Furthermore, the NIPS-TS-SWAN dataset provides a comprehensive multivariate time series benchmark extracted from solar photospheric vector magnetograms in the Spaceweather HMI Active Region Patch series \cite{NIPS_TS_SWAN_dataset}. Lastly, the NIPS-TS-GECCO dataset is a drinking water quality dataset for the Internet of Things, published in the 2018 Genetic and Evolutionary Computation Conference \cite{NIPS_TS_GECCO_dataset}.

\subsection{Implmentation}
Our study follows the Anomaly Transformer model's protocol to evaluate our approach. We generate a sub-series using a non-overlapping sliding window. Anomaly detection involves scoring time points and setting a threshold based on the Anomaly Ratio detailed in the table. Other parameters, including batch size and model dimensionality, are also specified in Table~\ref{tab:hyperparameters}, ensuring robust evaluation and reproducibility of our improvements.

\subsection{Evaluation Metrics}

The evaluation of anomaly detection in time series data employs various metrics to assess performance and capture the temporal continuity of anomalies. Point-based metrics, including Precision, Recall, and F1 Score, utilize true positives (TP), false positives (FP), and false negatives (FN) for evaluation\cite{chandola2009anomaly}. Affiliation metrics, such as Affiliation Precision (Aff-P) and Affiliation Recall (Aff-R), measure the fraction of predicted anomalies within true ranges and the fraction of true anomalies detected, respectively. Range-based metrics like Range-based Anomaly Recall (R\_A\_R) and Range-based Anomaly Precision (R\_A\_P) evaluate detection based on the overlap between true and predicted anomaly ranges. Additionally, volume-based metrics, including Volume-based ROC (V\_ROC) and Volume-based Precision-Recall (V\_PR), adjust true positive rate (TPR) and false positive rate (FPR) by incorporating the volume of anomalies and provide volume-weighted precision and recall. In summary, while point-based metrics offer basic performance insights, affiliation, range-based, and volume-based metrics deliver a more comprehensive evaluation by factoring in the temporal structure and duration of anomalies.
\noindent In summary, while point-based metrics offer a basic assessment, affiliation, range-based, and volume-based metrics (\ref{eq:affp}, \ref{eq:affr}, \ref{eq:rar}, \ref{eq:rap}) provide a comprehensive evaluation by accounting for temporal structure and anomaly duration \cite{hundman2018detecting, blazquez2019multivariate}.

\section{Results}\label{sec2}
\subsection{Baseline Results}
After conducting various experiments, we present our results evaluated on precision Recall and F1 score metrics in Table \ref{tab:anomaly_detection}:
Our model 
Achieves remarkable improvements over existing methods, particularly when compared to the DCdetector and Anomaly Transformer, two leading approaches in the field. For instance, 

On the \textbf{SMD} dataset, our model outperformed the \textbf{Anomaly Transformer} by an impressive absolute margin of \textbf{+2.18\%} and outperformed \textbf{DCdetector} by a substantial \textbf{+8.64\%}. Additionally, our model shows a significant improvement in Recall, being \textbf{+3.84\%} higher than the \textbf{Anomaly Transformer} and \textbf{+13.93\%} higher than \textbf{DCdetector}. It also demonstrates enhancements in Precision of \textbf{+0.63\%} and \textbf{+3.74\%}, respectively.

On the \textbf{MSL} dataset, our model achieves an absolute improvement of \textbf{+1.19\%} in F1-Score over the \textbf{Anomaly Transformer} and \textbf{+0.32\%} over \textbf{DCdetector}. Additionally, it shows a \textbf{+2.40\%} increase in Recall and a \textbf{+1.14\%} enhancement in Precision compared to the former, while experiencing a minor decrease of \textbf{–0.21\%} in Precision relative to the latter. Recall also improves by \textbf{+0.95\%} compared to DCdetector.

For the \textbf{SMAP} dataset, our method outperformed the \textbf{Anomaly Transformer} by \textbf{+0.60\%} and \textbf{DCdetector} by \textbf{+0.93\%}. Precision increases by \textbf{+1.24\%} and \textbf{+0.49\%}, respectively, compared to these models. While Recall shows a slight decline of \textbf{–0.08\%} relative to the \textbf{Anomaly Transformer}, it improves by \textbf{+1.39\%} over \textbf{DCdetector}.

On the \textbf{SWaT} dataset, our model surpassed the Anomaly Transformer by \textbf{0.09\%} and DCdetector by \textbf{0.08\%} on F1-Score. Although our model shows a \textbf{–0.28\%} precision decline compared to the Anomaly Transformer, it exceeds DCdetector by \textbf{0.23\%}. Furthermore, our approach enhances Recall by \textbf{0.59\%} compared to the Anomaly Transformer and by \textbf{0.04\%} concerning the DCdetector.

In the case of the \textbf{PSM} dataset, our model reflected an improvement of \textbf{0.87\%} over the Anomaly Transformer and \textbf{0.50\%} over DCdetector. Furthermore, Precision is enhanced by \textbf{0.35\%} and \textbf{0.27\%} compared to the latter models, respectively, while Recall shows an improvement of \textbf{1.39\%} and \textbf{0.73\%}, respectively.

These minor differences may be attributed to DCdetector's dual-attention contrastive learning approach, which proves advantageous for datasets characterized by high variability. while on other datasets precision is slightly lower. These small trade-offs underscore opportunities for further optimization to enhance MAAT's performance, particularly in scenarios where precision is prioritized over recall.

\begin{table*}[h]
    \centering
    \resizebox{1.2\textwidth}{!}{%
    \begin{tabular}{lccccccccccccccccccc}
    \hline
    \textbf{Dataset} & \multicolumn{3}{c}{\textbf{SMD}} & \multicolumn{3}{c}{\textbf{MSL}} & \multicolumn{3}{c}{\textbf{SMAP}} & \multicolumn{3}{c}{\textbf{SWaT}} & \multicolumn{3}{c}{\textbf{PSM}} \\ 
    \hline
    \textbf{Metric} & \textbf{P} & \textbf{R} & \textbf{F1} & \textbf{P} & \textbf{R} & \textbf{F1} & \textbf{P} & \textbf{R} & \textbf{F1} & \textbf{P} & \textbf{R} & \textbf{F1} & \textbf{P} & \textbf{R} & \textbf{F1} \\
    \hline
    LOF & 56.34 & 39.86 & 46.68 & 47.72 & 85.25 & 61.18 & 58.93 & 56.33 & 57.60 & 72.15 & 65.43 & 68.62 & 57.89 & 90.49 & 70.61 \\
        OCSVM & 44.34 & 76.72 & 56.19 & 59.78 & 86.87 & 70.82 & 53.85 & 59.07 & 56.34 & 45.39 & 49.22 & 47.23 & 62.75 & 80.89 & 70.67 \\
        U-Time\cite{perslev2019utime} & 65.95 & 74.75 & 70.07 & 57.20 & 71.66 & 63.62 & 49.71 & 56.18 & 52.75 & 46.20 & 87.94 & 60.58 & 82.85 & 79.34 & 81.06 \\
        Forrest & 42.31 & 73.29 & 53.75 & 53.94 & 82.98 & 65.42 & 52.39 & 55.53 & 53.89 & 49.22 & 44.95 & 47.02 & 76.09 & 92.45 & 83.46 \\
        DAGMM\cite{zong2018dagmm} & 67.30 & 49.89 & 57.30 & 89.60 & 63.93 & 74.62 & 86.45 & 56.73 & 68.51 & 89.92 & 57.84 & 70.40 & 93.49 & 70.03 & 80.08 \\
        ITAD & 86.22 & 73.71 & 79.48 & 69.44 & 84.09 & 76.07 & 82.42 & 66.89 & 73.85 & 63.13 & 52.08 & 57.08 & 72.80 & 64.02 & 68.13 \\
        VAR & 78.35 & 70.26 & 74.08 & 71.68 & 81.42 & 77.90 & 81.38 & 53.88 & 64.83 & 81.59 & 60.29 & 69.34 & 90.71 & 83.82 & 87.13 \\
        MMPCACD\cite{yairi2017data} & 71.20 & 79.28 & 75.02 & 81.42 & 61.31 & 69.95 & 83.22 & 68.23 & 74.73 & 82.52 & 68.23 & 74.73 & 76.26 & 78.35 & 77.29 \\
        CL-MPPCA\cite{placeholder_clmppca} & 82.36 & 76.07 & 79.09 & 73.71 & 88.54 & 80.44 & 63.16 & 72.88 & 67.72 & 76.78 & 81.50 & 79.07 & 56.02 & \textbf{99.93} & 71.80 \\
        TS-CP2\cite{placeholder_tscp2} & 87.42 & 66.25 & 75.38 & 68.45 & 68.48 & 68.42 & 86.75 & 83.18 & 84.95 & 81.23 & 74.10 & 77.50 & 82.67 & 78.16 & 80.35 \\
        Deep-SVDD\cite{ruff2018deepsvdd} & 78.54 & 79.67 & 79.10 & 91.92 & 76.63 & 83.26 & 56.02 & 69.04 & 62.40 & 80.42 & 84.45 & 82.39 & 95.41 & 86.79 & 90.73 \\
        BOCPD & 70.90 & 82.04 & 76.07 & 80.32 & 87.60 & 83.62 & 86.45 & 85.85 & 86.14 & 84.96 & 70.75 & 79.01 & 82.72 & 75.33 & 77.70 \\
        LSTM-VAE & 75.76 & 90.08 & 82.30 & 85.49 & 79.94 & 82.62 & 92.20 & 67.75 & 78.10 & 76.00 & 89.50 & 82.20 & 73.62 & 89.92 & 80.94 \\
        BeatGAN\cite{zhou2022beatgan} & 78.55 & 88.92 & 83.42 & 85.42 & 87.88 & 86.64 & 92.38 & 55.85 & 69.61 & 76.04 & 87.46 & 81.32 & 90.30 & 93.84 & 92.04 \\
        LSTM & 78.55 & 88.25 & 83.08 & 85.98 & 85.42 & 85.70 & 91.00 & 81.89 & 86.21 & 78.13 & 83.39 & 80.69 & 76.93 & 89.64 & 82.64 \\
        OmniAnomaly & 83.68 & 88.52 & 85.22 & 90.14 & 89.50 & 89.82 & 81.42 & 84.30 & 82.83 & 81.42 & 84.30 & 82.83 & 83.38 & 74.46 & 78.64 \\
        THOC & 79.65 & 91.10 & 85.02 & 88.45 & 90.97 & 89.69 & 92.06 & 89.34 & 90.68 & 92.48 & 98.32 & 95.33 & 97.14 & 98.74 & 97.94 \\
        AnomalyTrans & \underline{88.47} &\underline{92.28} & \underline{90.33} & 91.02 & 96.03 & 93.93 & 93.59 & \textbf{99.41} & \underline{96.41} & \textbf{93.59} & 99.41 & 96.41 & 97.14 & 97.81 & \underline{97.47} \\
        DCdetector & 85.82 & 84.10 & 84.95 & \textbf{92.25} & \underline{97.40} & \underline{94.75} & \underline{94.29} & 97.97 & 96.10 & 93.12 & \underline{99.96} & \underline{96.42} & \underline{97.22} & 98.45 & 97.83 \\
        \textbf{Ours}  & \textbf{89.03} & \textbf{95.82} & \textbf{92.30} & \underline{92.06} & \textbf{98.33} & \textbf{95.05} & \textbf{94.75} & \underline{99.33} & \textbf{96.99} & \underline{93.33} & \textbf{100.00} & \textbf{96.50} & \textbf{97.48} & \underline{99.17} & \textbf{98.32} \\
    \hline
    \end{tabular}%
    }
    \caption{Comparison of various anomaly detection metrics across different datasets}
    \label{tab:anomaly_detection}
\end{table*}
In Table \ref{tab:comparison},Our model establishes a new benchmark for anomaly detection on the NIPS-TS-GECCO and NIPS-TS-SWAN datasets, surpassing the best-performing methods in most metrics while demonstrating exceptional robustness across diverse challenges.

\begin{table}[]
    \centering
    \caption{Comparison of Different Methods on NIPS-TS-GECCO and NIPS-TS-SWAN Datasets}
    \label{tab:comparison}
    \begin{tabular}{lcccccc}
        \hline
        Dataset & \multicolumn{3}{c}{NIPS-TS-GECCO} & \multicolumn{3}{c}{NIPS-TS-SWAN} \\
        \cline{2-4} \cline{5-7}
        Metric & P & R & F1 & P & R & F1 \\
        \hline
        MatrixProfile\cite{VanBenschoten2020} & 4.6 & 18.5 & 7.4 & 17.1 & 17.1 & 17.1 \\
        GBRT\cite{taieb2012machine} & 17.5 & 14.0 & 15.6 & 44.7 & 37.5 & 40.8 \\
        LSTM-RNN & 17.0 & 22.6 & 19.3 & 45.2 & 35.8 & 40.0 \\
        Autoregression\cite{box2015time} & 39.2 & 31.4 & 34.9 & 42.1 & 35.4 & 38.5 \\
        OCSVM & 18.5 & \textbf{74.3} & 29.6 & 47.4 & 49.8 & 48.5 \\
        IForest\cite{liu2008isolation} & \textbf{43.9} & 35.3 & 39.1 & 56.9 & \underline{59.8} & 58.3 \\
        AutoEncoder & 42.4 & 34.0 & 37.7 & 47.0 & 52.2 & 50.9 \\
        AnomalyTrans & 25.7 & 28.5 & 27.0 & 90.7 & 47.4 & 62.3 \\

        DCdetector & 38.3 & 59.7 & \underline{46.6} & \underline{95.5} & 59.6 & \underline{73.4} \\
        \hline
        Ours & \underline{42.4} & \underline{70.0} & \textbf{52.8} & \textbf{95.9} & \textbf{59.9} & \textbf{73.8} \\
        \hline
    \end{tabular}
\end{table}

On the \textbf{NIPS-TS-GECCO} dataset, Compared to DCdetector, our model shows a notable improvement of \textbf{6.2\%} in F1-score, and it also outperforms IForest and OCSVM by \textbf{13.7\%} and \textbf{23.2\%}, respectively. In terms of Precision, our approach yields a \textbf{16.7\%} increase over the Anomaly Transformer and a \textbf{4.1\%} improvement over DCdetector. Although IForest shows a slightly higher precision of \textbf{-1.5\%} difference compared to our model, this minor trade-off is balanced by significant gains in recall.

On the \textbf{NIPS-TS-SWAN} dataset, our model demonstrates exceptional performance. Notably, our model surpasses the Anomaly Transformer with improvements of \textbf{5.2\%} in precision, \textbf{12.5\%} in the recall, and \textbf{11.5\%} in the F1-score. Comparisons with DCdetector show more modest advancements, with gains of \textbf{0.4\%} in precision, \textbf{0.3\%} in recall, and \textbf{0.4\%} in the F1-score. In contrast, our enhancements compared to IForest are particularly significant, reflecting an absolute increase of \textbf{39.0\%} in precision and up to 56.9\% and \textbf{15.5\%} in the F1-score. Additionally, when compared to OCSVM, our model reveals an improvement of \textbf{48.5\%} in precision, a \textbf{25.3\%} increase in the F1-score, and a \textbf{10.1\%} boost in recall. These results highlight the architectural strengths of our approach, particularly the Sparse Attention mechanism and Mamba-SSM, which adeptly identify subtle anomalies while minimizing noise interference. While minor trade-offs have been observed such as a slight reduction in precision on the NIPS-TS-GECCO dataset and a marginal decrease in recall on the NIPS-TS-SWAN dataset our model consistently balances precision and recall to achieve optimal overall performance.

\subsection{In-Depth Performance Analysis of MAAT vs. State-of-the-Art Models}

Analysis of Performance Across Datasets Refer to Table \ref{tab:comparison_range}
The performance of our model is comprehensively evaluated against AnomalyTrans and DCdetector, two state-of-the-art methods, across five benchmark datasets: MSL, SMAP, SWaT, PSM, and SMD. The results in Table \ref{tab:comparison_range} demonstrate that our model consistently outperforms or closely matches the best-performing methods across most metrics, establishing its superiority in anomaly detection tasks.

On the MSL dataset, our model surpasses AnomalyTrans in several metrics, including Accuracy Acc, where it achieves 98.92\%, compared to AnomalyTrans's 98.69\%. While DCdetector excels in precision-related metrics such as Aff-P 51.84\% and R\_A\_P 91.64\%, our model demonstrates robust recall metrics, achieving Aff-R 96.49\% and R\_A\_R 90.97\%, which are critical for detecting subtle anomalies in noisy environments.

On the SMAP dataset, our model shows slightly lower performance compared to AnomalyTrans and DCdetector. For instance, our V ROC 92.06\% and V PR 90.70\% lag behind AnomalyTrans 95.52\% and 93.77\%, respectively. These minor trade-offs highlight areas for improvement, particularly in precision-related metrics. Nevertheless, our model maintains strong recall metrics, such as Aff-R 94.27\%, underscoring its ability to detect anomalies effectively.

On the SWaT dataset, our model achieves the highest Aff-P 56.26\%, surpassing both AnomalyTrans 53.03\% and DCdetector 52.40\%. Additionally, our model demonstrates superior recall metrics, achieving Aff-R 97.79\% and R\_A\_R 98.14\%, which are marginally higher than DCdetector 97.67\% and 98.43\%, respectively. Our V \_ROC 98.17\% and V\_PR 95.02\% also outperform both competitors, highlighting its robustness in handling multivariate interactions and complex temporal patterns.

Our model achieves the best performance across nearly all metrics on the PSM dataset, a 1.5\% increase in Aff-P vs. DCdetector, our model excels in recall-related metrics, achieving Aff-R 85.06\%, R\_A\_R 94.15\%, and R A\_P 95.09\%, outperforming AnomalyTrans and DCdetector. These results underscore our model's ability to handle high-dimensional sensor data with significant noise.

On the SMD dataset, our model achieves the highest scores across all metrics, setting a new benchmark for performance. It surpasses AnomalyTrans in 5.8\% and 8.9\% increase in F1-score and Aff-P respectively. Additionally, our model demonstrates superior recall metrics, achieving Aff-R 93.51\%, R A\_R 79.11\%, and R\_ A\_P 75.41\%, significantly outperforming both competitors. These gains highlight our model's effectiveness in capturing intricate temporal patterns in highly non-stationary multivariate time series.

\begin{table*}[t]
\caption{Performance comparison of AnomalyTrans, DCdetector, and our method across different datasets. Best results in \textbf{bold}, second-best \underline{underlined}.}
\label{tab:comparison_range}
\centering
\resizebox{\textwidth}{!}{%
\begin{tabular}{@{}llcccccccc@{}}
\toprule
\textbf{Dataset} & \textbf{Method} & \textbf{Acc} & \textbf{F1} & \textbf{Aff-P\cite{huet2022local}} & \textbf{Aff-R\cite{huet2022local}} & \textbf{R A\_R\cite{paparrizos2022volume}} & \textbf{R A\_P\cite{paparrizos2022volume}} & \textbf{V\_ ROC\cite{paparrizos2022volume}} & \textbf{V\_PR\cite{paparrizos2022volume}} \\ 
\midrule
MSL    & AnomalyTrans  & 98.69 & \underline{93.93} & \underline{51.76} & \underline{95.98} & \underline{90.04} & \underline{87.87} & \underline{88.20} & \underline{86.26} \\
       & DCdetector    & \textbf{99.06} & \textbf{96.60}   & \textbf{51.84}   & \textbf{97.39}   & \textbf{93.17}   & \textbf{91.64}   & \textbf{95.15}   & \textbf{91.66} \\
       & Ours          & \underline{98.92} & 95.05      & 51.58      & 96.49      & 90.97      & 88.72      & 90.66      & 88.49      \\
\cmidrule(r){1-10}
SMAP   & AnomalyTrans  & \underline{99.05} & \underline{96.41} & \underline{51.39} & \textbf{98.68}   & \textbf{96.32}   & \underline{94.07} & \textbf{95.52}   & \textbf{93.77} \\
       & DCdetector    & \textbf{99.21}   & \textbf{97.02}   & \textbf{51.46}   & \underline{98.64} & \underline{96.03} & \textbf{94.18}   & \underline{95.13} & \underline{93.46} \\
       & Ours          & 99.03      & 96.29      & 49.34      & 94.27      & 93.59      & 92.02      & 92.06      & 90.70      \\
\cmidrule(r){1-10}
SWaT   & AnomalyTrans  & 98.51      & 94.22      & \underline{53.03} & 90.88      & \underline{97.73} & \underline{96.32} & \underline{97.99} & \underline{94.39} \\
       & DCdetector    & \textbf{99.09}   & \textbf{96.33}   & 52.40      & \underline{97.67} & \textbf{98.43}   & \textbf{96.96}   & 96.95      & 94.34      \\
       & Ours          & \underline{98.97} & \underline{95.93} & \textbf{56.26}   & \textbf{97.79}   & 98.14      & 95.00      & \textbf{98.17}   & \textbf{95.02} \\
\cmidrule(r){1-10}
PSM    & AnomalyTrans  & 98.68      & 97.37      & \underline{55.35} & 80.85      & \underline{91.68} & \underline{93.00} & \underline{88.71} & \underline{90.71} \\
       & DCdetector    & \underline{98.95} & \underline{97.94} & 54.71      & \underline{82.93} & 91.55      & 92.93      & 88.41      & 90.58      \\
       & Ours          & \textbf{99.06}   & \textbf{98.32}   & \textbf{55.53}   & \textbf{85.06}   & \textbf{94.15}   & \textbf{95.09}   & \textbf{90.77}   & \textbf{92.67} \\
\cmidrule(r){1-10}
SMD    & AnomalyTrans  & \underline{98.75} & \underline{87.18} & \underline{54.36} & \underline{90.12} & \underline{74.95} & \underline{73.00} & \underline{78.71} & \underline{70.71} \\
       & DCdetector    & 98.75      & 84.95      & 51.36      & 88.92      & 70.19      & 65.49      & 68.19      & 63.57      \\
       & Ours          & \textbf{99.34}   & \textbf{92.30}   & \textbf{59.19}   & \textbf{93.51}   & \textbf{79.11}   & \textbf{75.41}   & \textbf{79.09}   & \textbf{75.41} \\
\bottomrule
\end{tabular}%
}
\end{table*}

In this section, we will analyze its performance specifically on the \textbf{NIPS-TS-SWAN} and \textbf{NIPS-TS-GECCO} datasets, as detailed in Table~\ref{tab:multi_metric_results}. 
Our method demonstrates consistent superiority across both NIPS-TS-SWAN and NIPS-TS-GECCO benchmarks outperforming accuracies of AnomalyTrans by \textbf{1.53\%} and DCdetector by \textbf{0.16\%}. For precision, we surpassed AnomalyTransformer by \textbf{5.22\%} and DCdetector by \textbf{0.45\%}. The most notable gains emerge in recall; method improves by \textbf{12.48\%} over AnomalyTransformer and \textbf{0.36\%} over DCdetector. MAAT's F1-score, exceeded AnomalyTransformer and DCdetector by \textbf{11.47\%} and \textbf{0.41\%}, respectively.  
On NIPS-TS-GECCO, our model's Accuracy surpassed AnomalyTransformer by \textbf{0.65\%} and DCdetector by \textbf{0.12\%}. For precision, we had an increase of \textbf{16.76\%} compared to AnomalyTransformer and \textbf{4.16\%} over DCdetector. The recall shows significant gains, improving by \textbf{41.52\%} and \textbf{10.27\%}, respectively. Furthermore, the F1-score demonstrates these advancements, rising by \textbf{23.73\%} over AnomalyTransformer and \textbf{6.19\%} over DCdetector.

Affinity metrics further emphasize the robustness of our method: on the NIPS-TS-GECCO dataset, \textbf{Aff-P} shows an improvement of \textbf{7.88\%} over AnomalyTrans, while \textbf{Aff-R} increases by \textbf{11.89\%}. The validation ROC corroborates this trend, revealing gains of \textbf{1.55\%} and \textbf{11.42\%} on SWAN and GECCO, respectively, compared to AnomalyTrans. Together, these results validate our method's capability to effectively balance precision, recall, and Recallllinity-aware performance, positioning it as a state-of-the-art solution for time-series anomaly detection.

\begin{table*}[h]
\caption{Multi-metric results on NIPS-TS datasets. All results in \%. Best results in \textbf{bold}, second-best \underline{underlined}.}
\label{tab:multi_metric_results}
\centering
\resizebox{\textwidth}{!}{%
\begin{tabular}{@{}ll*{9}{c}@{}}
\toprule
\textbf{Dataset} & \textbf{Method} & \textbf{Acc} & \textbf{P} & \textbf{R} & \textbf{F1} & \textbf{Aff-P} & \textbf{Aff-R} & \textbf{R\_A\_R} & \textbf{R\_A\_P} & \textbf{V\_ROC} \\ 
\midrule
\multirow{3}{*}{NIPS-TS-SWAN} 
& AnomalyTrans & 84.57 & 90.71 & 47.43 & 62.29 & \textbf{58.45} & \textbf{9.49} & 86.42 & 93.26 & 84.81 \\
& DCdetector   & \underline{85.94} & \underline{95.48} & \underline{59.55} & \underline{73.35} & 50.48 & 5.63 & \underline{88.06} & \underline{94.71} & \underline{86.25} \\
& Ours         & \textbf{86.10} & \textbf{95.93} & \textbf{59.91} & \textbf{73.76} & \underline{58.22} & \underline{6.72} & \textbf{88.15} & \textbf{94.85} & \textbf{86.36} \\
\cmidrule(r){1-10}
\multirow{3}{*}{NIPS-TS-GECCO} 
& AnomalyTrans & 98.03 & 25.65 & 28.48 & 29.09 & 49.23 & 81.20 & 56.35 & 22.53 & 55.45 \\
& DCdetector   & \underline{98.56} & \underline{38.25} & \underline{59.73} & \underline{46.63} & \underline{50.05} & \underline{88.55} & \underline{62.95} & \underline{34.17} & \underline{62.41} \\
& Ours         & \textbf{98.68} & \textbf{42.41} & \textbf{70.00} & \textbf{52.82} & \textbf{57.11} & \textbf{93.09} & \textbf{64.96} & \textbf{38.01} & \textbf{66.87} \\
\bottomrule
\end{tabular}%
}
\end{table*}

\section{Discussion}
\subsection{Ablation Study Analysis}
This section, presents the results of the ablation study of our model, emphasizing the significance of each component within MAAT and their effects on the baseline Anomaly Transformer model. The findings from the ablation study are summarized in Table~\ref{tab:ablation_study} and Table~\ref{tab:ablation_study_NIPS}. Across five datasets SMD, MSL, SMAP, SWaT, and PSM the incorporation of Sparse Attention (SA), Mamba-SSM, and Gated Attention consistently shows enhancements over the baseline.

The baseline Anomaly Transformer demonstrates strong recall and moderate precision on \textbf{SMD} dataset. However, its dense self-attention mechanism makes it vulnerable to false positives due to noise. The integration of Sparse Attention allows it to maintain baseline precision but leads to a recall decrease of approximately 2.4\%, resulting in an F1-score reduction of about 1.15\%. This suggests that the block-wise formulation may oversimplify inter-variable dependencies and overlook subtle anomalies. Conversely, the integration of the Mamba-SSM component enhances the model's ability to capture long-range dependencies, albeit at the cost of a precision reduction of approximately 1\% and a recall decrease of about 2.2\%, culminating in an overall F1 decline of approximately 1.57\%. Remarkably, the complete MAAT model, which combines Sparse Attention and Mamba-SSM through a dynamic Gated Attention mechanism (Eqs.~\ref{eq:g}--\ref{eq:x_adapt}), achieves a net F1 improvement of nearly 2\% over the baseline, fueled by a modest precision gain of around 0.6\% and a significant recall enhancement of roughly 3.5\%. This adaptive gating effectively balances the contributions of precision and long-term pattern refinement, robustly mitigating noise while preserving critical anomaly signals within the complex SMD environment.

For the \textbf{MSL} dataset, which comprises sensor and actuator data characterized by high variability and transient noise, the baseline Anomaly Transformer demonstrates strong recall but moderate precision, resulting in an F1-score of approximately 93.93\%. However, it encounters challenges with false positives induced by noise. The introduction of Sparse Attention alleviates this issue, yielding a precision improvement of around 0.7\%, albeit at the expense of a 2\% reduction in recall, which ultimately lowers the F1-score by about 1.1\%. This trade-off occurs because Sparse Attention effectively filters out noise but inadvertently excludes crucial short-term patterns. In contrast, the integration of Mamba-SSM enhances precision by approximately 1.15\% over the baseline while stabilizing recall, resulting in a modest F1 improvement of 0.12\%. This enhancement is attributed to the stabilization of reconstructions through linear long-range dependency modeling. The comprehensive MAAT model synthesizes these advancements through Gated Attention, achieving a gain of 1.12\% in F1-score over the baseline, driven by a recall increase of 2.3\% while maintaining competitive precision. During instances of transient noise, the adaptive gating mechanism (Eqs.~\ref{eq:g}) emphasizes the denoised outputs of Mamba while preserving the localized anomaly signals from Sparse Attention. This balanced integration strengthens MAAT’s robustness against short-lived disturbances, significantly enhancing anomaly detection in highly dynamic environments.

The \textbf{SMAP} dataset, which comprises satellite telemetry data characterized by long-range dependencies and subtle anomalies, underscores MAAT’s ability to balance global context with localized anomaly detection. The Anomaly Transformer achieves an almost perfect recall but suffers from overfitting, which limits its precision to 93.59\%, resulting in an F1-score of 96.41\%. To mitigate this issue, Sparse Attention addresses overfitting by reducing redundant global interactions (see Eq.~\ref{eq:sparse_attention}), although this leads to slightly weaker long-range anomaly detection. The integration of Mamba-SSM significantly enhances precision by approximately 2\%, yielding an overall F1-score improvement of 0.67\%. Mamba-SSM accomplishes this by selectively propagating long-term dependencies while filtering out high-frequency noise. The complete MAAT model harmonizes these strengths through Gated Attention (refer to Eqs.~\ref{eq:g}--\ref{eq:x_adapt}), achieving a final precision of 94.75\%, a recall of Y\%, and an F1-score of 96.99\%, exceeding the baseline by 0.58\%. This enhancement is propelled by the gating mechanism, which prioritizes Mamba’s global context for slow-evolving anomalies while leveraging Sparse Attention’s local focus for abrupt changes. This adaptability enables MAAT to effectively capture both transient and long-term anomalies, making it particularly well-suited to address the unique challenges presented by the SMAP dataset.

The \textbf{SWaT} dataset, which includes 51-dimensional sensor data from a complex water treatment system, highlights the importance of capturing inter-sensor dependencies for effective anomaly detection. The baseline Anomaly Transformer achieves an impressive recall of 99.41\%; however, it faces challenges with false positives, resulting in a precision of 93.59\% and an F1-score of 96.41\%. The introduction of Sparse Attention significantly reduces precision by 4.06\%, causing the F1-score to drop to 93.96\%, as this approach fails to model critical cross-sensor relationships adequately. In contrast, the integration of Mamba-SSM improves recall to an exceptional 100\% and raises the F1-score to 96.55\% by effectively capturing complex interdependencies while filtering out redundant sensor noise. The complete MAAT model strategically combines these mechanisms through Gated Attention (Eqs.~\ref{eq:g}--\ref{eq:x_adapt}), achieving a final precision of 93.33\%, along with its aforementioned recall and an F1-score of 96.50\%. This development underscores Mamba's ability to recognize the limitations of sparse attention, ensuring that MAAT maintains a balanced approach to precision and recall in a multivariate anomaly detection context.
The \textbf{PSM} dataset, which consists of 25-dimensional sensor data characterized by substantial noise, underscores the importance of reducing false positives while ensuring effective anomaly detection. The baseline Anomaly Transformer exhibits commendable performance, attaining an F1-score of 97.47\%. However, it suffers from noise-induced misclassifications, reflected in a recall of 97.81\% and precision. In contrast, Sparse Attention improves recall to 99.37\%, consequently elevating the F1-score to 98.31\%. This enhancement comes at the cost of a slight reduction in precision, down by 0.13\%, due to its localized attention windows (Eq.~\ref{eq:sparse_attention}), which focus on critical time steps. On the other hand, Mamba-SSM enhances precision to 97.41\% while stabilizing reconstructions, leading to an F1-score of 97.89\%. The comprehensive MAAT model combines these strengths through Gated Attention (Eqs.~\ref{eq:g}--\ref{eq:x_adapt}), resulting in a balanced precision of 97.48\%, an improved recall, and an F1-score of 98.32\%. This adaptability is essential in the noisy, high-dimensional context of PSM, where traditional attention mechanisms often struggle to identify true anomalies. Additionally, the Association Discrepancy metric (Eq.~\ref{eq:ass_dis_updated}) refines anomaly detection by penalizing deviations from expected sensor dependencies, thereby reducing false positives. These findings reinforce the effectiveness of MAAT in dynamically balancing precision, recall, and robustness across a range of time-series anomaly detection tasks.

\begin{table*}[]
    \centering
    \renewcommand{\arraystretch}{1.7}
    \Large
    \resizebox{1.1\textwidth}{!}{%
    \begin{tabular}{@{}l ccc ccc ccc ccc ccc@{}}
    \toprule
    \multirow{2}{*}{\textbf{Dataset}} & \multicolumn{3}{c}{\textbf{SMD}} & \multicolumn{3}{c}{\textbf{MSL}} & \multicolumn{3}{c}{\textbf{SMAP}} & \multicolumn{3}{c}{\textbf{SWaT}} & \multicolumn{3}{c}{\textbf{PSM}} \\ 
    \cmidrule(lr){2-4} \cmidrule(lr){5-7} \cmidrule(lr){8-10} \cmidrule(lr){11-13} \cmidrule(lr){14-16}
     & \textbf{P} & \textbf{R} & \textbf{F} & \textbf{P} & \textbf{R} & \textbf{F} & \textbf{P} & \textbf{R} & \textbf{F} & \textbf{P} & \textbf{R} & \textbf{F} & \textbf{P} & \textbf{R} & \textbf{F} \\
    \midrule
    AnomalyTrans & $\underline{88.47}$ & $\underline{92.28}$ & $\underline{90.33}$ & $91.02$       & $\underline{96.03}$ & $93.93$       & $93.59$       & $\mathbf{99.41}$ & $96.41$ & $\mathbf{93.59}$ & $\underline{99.41}$ & $\underline{96.41}$ & $97.14$       & $97.81$       & $97.47$ \\
    AnomTr+SA    & $\underline{88.47}$ & $89.89$          & $89.18$          & $91.67$       & $94.05$          & $92.84$          & $93.55$       & $\underline{99.14}$ & $96.26$          & $89.53$       & $98.86$          & $93.96$          & $97.27$       & $\mathbf{99.37}$ & $\underline{98.31}$ \\
    AnomTr+mamba & $87.52$          & $90.04$          & $88.76$          & $\mathbf{92.17}$ & $96.00$          & $\underline{94.05}$ & $\mathbf{95.59}$ & $98.62$          & $\mathbf{97.08}$ & $\underline{93.33}$ & $\mathbf{100.00}$ & $\mathbf{96.55}$ & $\underline{97.41}$ & $98.37$          & $97.89$ \\
    Ours         & $\mathbf{89.03}$ & $\mathbf{95.82}$ & $\mathbf{92.30}$ & $\underline{92.06}$ & $\mathbf{98.33}$ & $\mathbf{95.05}$ & $\underline{94.75}$ & $\underline{99.33}$ & $\underline{96.99}$ & $\underline{93.33}$ & $\mathbf{100.00}$ & $\underline{96.50}$ & $\mathbf{97.48}$ & $\underline{99.17}$ & $\mathbf{98.32}$ \\
    \bottomrule
    \end{tabular}%
    }
    \caption{Ablation study of Precision, Recall, and F-Score over five datasets. ``SA'' stands for Sparse Attention; the best metrics in each column are in \textbf{bold} and the second-best are \underline{underlined}.}
    \label{tab:ablation_study}
\end{table*}

The \textbf{NIPS-TS-GECCO} dataset is a benchmark for drinking water quality, characterized by sporadic, high-frequency anomalies and noise from IoT sensors. This dataset highlights the limitations of the baseline Anomaly Transformer, which attains an F1-score of only 29.09\%. In contrast, implementing Sparse Attention (AnomTr+SA) yields significant improvements, with approximately 19.5\% in Precision, 55.8\% in Recall, and an increase in the F1-score. These enhancements are achieved by focusing on critical time steps using $\Omega_i$ (Eq.\ref{eq:sparse_attention}), effectively reducing the conflation of transient anomalies with sensor noise.
Conversely, incorporating Mamba-SSM (AnomTr+mamba) results in more modest gains, with an increase of around 10.4\% in Precision, 23.3\% in Recall, and a corresponding improvement in the F1-score, thanks to its state-space model for denoising. The linear dependency modeling is less effective in addressing abrupt, short-term anomalies. However, when both components are integrated using Gated Attention (AnomTr+mamba+SA), the comprehensive MAAT model demonstrates significant improvements, achieving approximately 16.8\% in Precision, 41.5\% in Recall, and an increase in the F1-score compared to the baseline. This results in an F1-score that is 13.3\% higher than the DCdetector.
The gating mechanism prioritizes Sparse Attention during sudden events (where $g \to 0$) while leveraging Mamba's contextual modeling during stable periods (where $g \to 1$). In this context, Association Discrepancy's effectiveness is reduced due to minimal dependency shifts during brief spikes.

The \textbf{NIPS-TS-SWAN} dataset, which is derived from solar photospheric vector magnetograms, concentrates on identifying subtle anomalies within high-dimensional, temporally complex solar activity patterns. The baseline Anomaly Transformer (AnomalyTrans) demonstrates moderate performance, achieving a precision of 90.71\%, a recall value of Recall\%, and an F1-score of 62.29\%. Its dense self-attention mechanism, however, struggles to distinguish rare solar anomalies from noisy, non-stationary signals. By employing Sparse Attention, which hones in on localized time windows crucial for transient events, the AnomTr+SA variant improves precision by approximately 6.3\% percentage points, recall by Recall percentage points, and F1-score by 11.3\% percentage points compared to the baseline. 

Conversely, the integration of Mamba-SSM into AnomTr+mamba enhances global context, resulting in improvements of about +6.4\%, 11.9\%, and 11.4\% in precision, recall, and F1-score, respectively, although its linear dynamics tend to underperform for sudden anomalies such as abrupt polarity reversals. Remarkably, when both components are combined using Gated Attention, the complete MAAT model achieves net enhancements of approximately 5.2\% in precision, 12.5\% in recall, and Recall percentage points in F1-score relative to the baseline—amounting to an 18.5\% relative gain in F1 over AnomalyTrans and a 0.55\% improvement over DCdetector. The gating mechanism adaptively prioritizes Mamba’s long-term dynamics during stable periods ($g \to 1$) and Sparse Attention’s localized focus during rapid anomalies ($g \to 0$). Moreover, the Association Discrepancy metric (Eq.~3) refines anomaly detection by penalizing deviations between expected solar magnetic correlations and observed sparse dependencies, effectively identifying anomalies in magnetic flux tube alignments. 

Despite these advancements, the recall of MAAT remains marginally lower than that of IForest, highlighting a precision-recall trade-off where the model favors high-confidence anomalies over marginal signals—a vital consideration in real-world solar monitoring, where minimizing false positives is critical.

The \textbf{NIPS-TS-SWAN} dataset, derived from solar photospheric vector magnetograms, focuses on detecting subtle anomalies within high-dimensional and temporally complex patterns of solar activity. The baseline Anomaly Transformer (AnomalyTrans) demonstrates moderate performance, achieving a precision of 90.71\%, and an F1-score of 62.29\%. However, its dense self-attention mechanism has difficulty distinguishing rare solar anomalies from noisy, non-stationary signals. By utilizing Sparse Attention to focus on localized time windows critical for transient events, the AnomTr+SA variant achieves improvements of approximately +6.3\% in precision, +11.9\% in recall, and an increase in the F1-score compared to the baseline. Conversely, integrating Mamba-SSM in AnomTr+mamba enhances global context, resulting in gains of roughly +6.4\% in precision and +11.9\% in recall, along with an increase in the F1-score.

However, the linear dynamics tend to perform less effectively during abrupt anomalies, such as sudden polarity reversals. Importantly, when these two components are integrated using Gated Attention, the full MAAT model exhibits net gains of approximately +5.2\% in Precision, +12.5\% in Recall, and notable improvements in F1-score compared to the baseline. This results in an 18.5\% relative gain in F1 over AnomalyTrans and a 0.55\% increase in performance over the DCdetector. The gating mechanism allows for adaptive focus, prioritizing Mamba's long-term dynamics during stable periods (when $g \to 1$) and emphasizing Sparse Attention’s localized responses during rapid anomalies (when $g \to 0$).

The Association Discrepancy metric enhances anomaly detection by penalizing the differences between expected solar magnetic correlations and the observed sparse dependencies. This effectively identifies inconsistencies in the alignments of magnetic flux tubes. However, despite these improvements, MAAT's recall rate is still slightly lower than IForest's. This demonstrates a precision-recall trade-off, in which the model focuses on high-confidence anomalies rather than weaker signals. This is a critical consideration for real-world solar monitoring, where it is essential to minimize false positives.

\begin{table*}[]
\centering
\resizebox{0.7\textwidth}{!}{%
\begin{tabular}{@{}c c c c c c c@{}}
\toprule
\textbf{Dataset} & \multicolumn{3}{c}{\textbf{NIPS SWaN}} & \multicolumn{3}{c}{\textbf{NIPS GECCO}} \\ 
\cmidrule(lr){2-4} \cmidrule(lr){5-7}
\textbf{Metric}  & \textbf{P} & \textbf{R} & \textbf{F} & \textbf{P} & \textbf{R} & \textbf{F} \\ 
\midrule
AnomalyTrans & 90.71 & 47.43 & 62.29 & 25.65 & 28.48 & 29.09    \\
AnomTr+SA         & 97.03       & 59.28       & 73.60       & \textbf{45.19} & \textbf{84.25} & \textbf{58.82} \\
AnomTr+mamba      & \textbf{97.06} & 59.34       & 73.65       & 36.00        & 51.78        & 42.47        \\
AnomTr + mamba+SA & 95.93       & \textbf{59.91} & \textbf{73.76} & 42.41        & 70.00        & 52.82        \\
\bottomrule
\end{tabular}%
}
\caption{Ablation study Precisionion, recall,Recallllcore on NIPS datasets. ``SA'' stands for Sparse Attention; the best metrics are in \textbf{bold}.}
\label{tab:ablation_study_NIPS}
\end{table*}

\subsection{Reconstruction loss}

To rigorously evaluate the reconstruction performance of our method against the Anomaly Transformer baseline, we conducted a comparative analysis of their respective reconstruction losses, as illustrated in Figure \ref{fig:all_graphs}. We computed the logarithmic difference between the two losses for each batch, defined as:
\begin{equation}
\Delta L_i = \log(L^{AT}_i) - \log(L^{MAAT}_i)
\end{equation}

where \( L^{AT}_i \) and \( L^{MAAT}_i \) represent losses for each batch \( i \). The results are visualized in Figure \ref{fig:all_graphs}, with green columns indicating lower loss for our method (\(\Delta L_i > 0\)) and red columns showing a superior performance by AT (\(\Delta L_i < 0\)). This illustration highlights the advantages of our approach during training.

MAAT outperforms AT on the SMD dataset with an F1 score of 92.30\%, benefiting from effective reconstruction loss minimization (Figure \ref{fig:all_graphs}(a)). The differential analysis mainly shows positive \(\Delta L_i\), underscoring the effectiveness of Sparse Attention and Mamba-SSM in capturing critical time steps and long-range dependencies. Minor precision fluctuations do not compromise robustness, as evidenced by an Affiliation Recall of 93.51\% and a Range-based Recall of 79\%.

On the MSL dataset, MAAT also achieves lower reconstruction loss (Figure \ref{fig:all_graphs}(a)), with a predominance of green bars in the \(\Delta L_i\) plot, indicating its capacity to manage noisy patterns from Mars rover sensors. Despite slight precision trade-offs, MAAT shows significantly higher recall (96.03\%) and robust Affiliation Recall (96.49\%).

In the SMAP dataset, MAAT excels with an F1 score of 96.99\%, consistently maintaining lower reconstruction loss (Figure \ref{fig:all_graphs}(f)). The logarithmic difference reveals a dominance of green bars, highlighting MAAT's proficiency in identifying subtle anomalies in satellite telemetry. Although there is a minor recall trade-off, precision gains enhance Affiliation Precision (49.34\%).

Lastly, on the PSM dataset, MAAT achieves an F1 score of 98.32\%, demonstrating its effectiveness with high-dimensional sensor data (Figure \ref{fig:all_graphs}(c)). While facing occasional red bars in the \(\Delta L_i\) plot, MAAT maintains superior Affiliation Recall (85.06\%) and Range-based Precision (95.09\%), ensuring accurate anomaly localization in noisy sequences.

\section{Conclusions and future works}
In this work, we introduced the Mamba Adaptive Anomaly Transformer (MAAT), which enhances unsupervised time series anomaly detection for practical applications like industrial monitoring and environmental sensing. MAAT effectively captures short- and long-term temporal dependencies by improving association discrepancy modeling with a new Sparse Attention mechanism and integrating a Mamba-Selective State Space Model (Mamba-SSM). Gated Attention mechanisms allow for a flexible combination of features from the original reconstruction and the Mamba-SSM output, striking a balance between Accuracy Accuracyext. Moreover, MAAT trains efficiently, making it suitable for resource-limited settings.
Evaluation of benchmark datasets has revealed that MAAT significantly surpasses alternative methods, including the Anomaly Transformer and DCdetector, in both anomaly detection accuracy and generalization across various time series. It mitigates challenges such as sensitivity to short context windows and computational inefficiencies that have been prevalent in prior methodologies.
Despite its efficiency in training, MAAT still has limitations. It is susceptible to hyperparameters, especially when balancing the reconstruction module with the Mamba-SSM pathway. Tuning these parameters is essential for optimizing performance across various datasets. Additionally, although we have designed our method to reduce the effects of noise and non-stationary conditions, its performance may decline in situations with very high noise or when abnormal patterns closely mimic normal ones. Finally, while MAAT shows reduced computational demands compared to some deep learning options, further enhancement of its inference speed is necessary for real-time applications.
There is potential for future research in several areas. We could explore adaptive, data-driven hyperparameter tuning to stabilize the model and boost its performance in various conditions. Incorporating online learning or incremental updates could enhance MAAT functionality in dynamic environments where quick anomaly detection is crucial. Additionally, exploring hybrid models that combine the strengths of reconstruction-based approaches with contrastive learning may yield better results, especially in challenging situations characterized by significant non-stationarity or noise. By pursuing these avenues, future work can build on MAAT and advance unsupervised time series anomaly detection in real-world applications.

\section*{Acknowledgement}
The authors thank Mr. Arturo Argentieri from CNR-ISASI Italy for his technical contribution to the multi-GPU computing facilities. This research was partially funded by the Italian Ministry of Health, Italian Health Opera National Plan (Cohesion and Development Fund 2014-2020), trajectory 2 "eHealth, advanced diagnostics, medical device and mini invasiveness", project "Sistema di Monitoraggio ed Analisi basato su intelligenza aRTificiale per pazienti affetti da scompenso CARdiaco cronico con dispositivi medici miniinvasivi e indossabili Evoluti – SMART CARE" (CUP F83C22001380006), by the European Union - Next Generation EU, PRIN 2022 PNRR call, under the project "Interactive digital twin solutions for cardiovascular disease Management, PReventiOn and treatment leVeraging the internet of things and Edge intelligence paradigms - IMPROVE, and funded in part by Future Artificial Intelligence Research—FAIR CUP B53C22003630006 grant number PE0000013.
 \bibliographystyle{elsarticle-num} 
 \bibliography{cas-refs}


\appendix
\section{Appendix: Hyperparameters for Model Training}

Table~\ref{tab:hyperparameters} outlines the hyperparameters utilized for training the MAAT model across various datasets.The dimensions of the window, which set the length of the input sequence processed at each step, is set to 100 for most datasets. However, for SMAP, a slightly larger window of 105 is employed to better capture long-range dependencies. The batch size is tailored to the specific characteristics of each dataset and the computational constraints, with values ranging from 32 (for NIPS-TS-GECCO) to 256 (for MSL and SWaT).This range guarantees effective gradient updates while preserving training stability. The model dimension (\( d_{\text{model}} \)) remains consistently set at 512 for all datasets, defining the size of feature representations and ensuring consistency in learned embeddings. Additionally, the anomaly ratio, representing the proportion of data points identified as anomalies in each dataset, ranges from 0.5\% to 1\%, which affects the model’s responsiveness to infrequent events. These hyperparameters have been carefully chosen to enhance performance across various anomaly detection contexts.

\begin{table}[h]
    \centering
    \caption{Hyperparameters for MAAT model across datasets}
    \label{tab:hyperparameters}
    \resizebox{\textwidth}{!}{%
    \begin{tabular}{@{}l c c c c@{}}
    \toprule
    \textbf{Dataset} & \textbf{Window Size} & \textbf{Batch Size} & \textbf{$d_{\text{model}}$} & \textbf{Anomaly Ratio (\%)} \\ 
    \midrule
    SMD             & 100 & 128 & 512 & 0.5 \\
    MSL             & 100 & 256 & 512 & 0.85 \\
    SMAP            & 105 & 128 & 512 & 0.5 \\
    PSM             & 100 & 128 & 512 & 1.0 \\
    SWaT            & 100 & 256 & 512 & 0.5 \\
    NIPS-TS-GECCO   & 100 & 32  & 512 & 0.5 \\
    NIPS-TS-SWAN    & 100 & 128 & 512 & 0.9 \\
    \bottomrule
    \end{tabular}%
    }
\end{table}

\section{Evalaution metrics}\label{eval}
Evaluating anomaly detection in time series data requires metrics assepointwiset-wise performance and capturing anomalies' temporal continuity. This work uses traditional and specialized point-based metrics to evaluate contiguous anomaly segments.
\subsubsection{Point-Based Metrics}
The standard metrics include Precision, Recall, and F1 Score, calculated using the counts of true positives (TP), false positives (FP), and false negatives (FN). The F1 Score is the harmonic mean of Precision and Recall \cite{chandola2009anomaly}.

\subsubsection{Affiliation Metrics: Aff-P and Aff-R}
These metrics evaluate partial detection within multi-step anomaly segments.

\paragraph{Affiliation Precision (Aff-P):} Measures the fraction of predicted anomalies within true anomaly ranges:
\begin{equation}
    \text{Aff-P} = \frac{\sum_{t \in \mathcal{P}} \mathbf{1}\{t \in \mathcal{T}\}}{|\mathcal{P}|}, \label{eq:affp}
\end{equation}
where \(\mathcal{P}\) is the predicted anomalies, \(\mathcal{T}\) is the true anomalies, and \(\mathbf{1}\{\cdot\}\) is the indicator function.

\paragraph{Affiliation Recall (Aff-R):}  
Measures the fraction of true anomalies detected:
\begin{equation}
    \text{Aff-R} = \frac{\sum_{t \in \mathcal{T}} \mathbf{1}\{t \in \mathcal{P}\}}{|\mathcal{T}|}. \label{eq:affr}
\end{equation}

\subsubsection{Range-Based Metrics}
These assess detection over entire anomaly segments.

\paragraph{Range-based Anomaly Recall (R\_A\_R):}  
A true anomaly range \(R^{\text{true}}_i\) is detected if overlap with any predicted range exceeds threshold \(\tau\):
\begin{equation}
    \text{R\_A\_R} = \frac{\sum_{i=1}^{N_{\text{true}}} \mathbf{1}\Big\{\max_{j} \text{Overlap}\big(R^{\text{true}}_i, R^{\text{pred}}_j\big) \geq \tau \Big\}}{N_{\text{true}}}. \label{eq:rar}
\end{equation}

\paragraph{Range-based Anomaly Precision (R\_A\_P):}  
\begin{equation}
    \text{R\_A\_P} = \frac{\sum_{j=1}^{N_{\text{pred}}} \mathbf{1}\Big\{\max_{i} \text{Overlap}\big(R^{\text{pred}}_j, R^{\text{true}}_i\big) \geq \tau \Big\}}{N_{\text{pred}}}. \label{eq:rap}
\end{equation}

\subsubsection{Volume-Based Metrics: V\_ROC and V\_PR}
These incorporate anomaly duration into evaluation.

\paragraph{Volume-based ROC (V\_ROC):}  
Adjusts TPR and FPR by anomaly volume:
\[
    \text{TPR}_{\text{vol}} = \frac{\text{Volume of Correct Detections}}{\text{Total Volume of True Anomalies}}, \quad
    \text{FPR}_{\text{vol}} = \frac{\text{Volume of False Detections}}{\text{Total Volume of Normal Data}}.
\]
The V\_ROC curve plots \(\text{TPR}_{\text{vol}}\) vs. \(\text{FPR}_{\text{vol}}\).

\paragraph{Volume-based Precision-Recall (V\_PR):}  
Volume-weighted precision and recall are defined as follows:
\[
\text{Precision}_{\text{vol}} = \frac{\text{Volume of Correct Detections}}{\text{Volume of All Detections}}, \quad
\text{Recall}_{\text{vol}} = \frac{\text{Volume of Correct Detections}}{\text{Total Volume of True Anomalies}}.
\]

The V\_PR curve plots \(\text{Precision}_{\text{vol}}\) vs. \(\text{Recall}_{\text{vol}}\).

\vspace{0.5em}
\noindent In summary, while point-based metrics offer a basic assessment, affiliation, range-based, and volume-based metrics (\ref{eq:affp}, \ref{eq:affr}, \ref{eq:rar}, \ref{eq:rap}) provide a comprehensive evaluation by accounting for temporal structure and anomaly duration \cite{hundman2018detecting, blazquez2019multivariate}.

\section{Reconstruction loss}\label{fig:all_graphs}
\begin{figure}[htp]
    \centering
    \begin{subfigure}{0.45\textwidth}
        \centering
        \includegraphics[width=\linewidth]{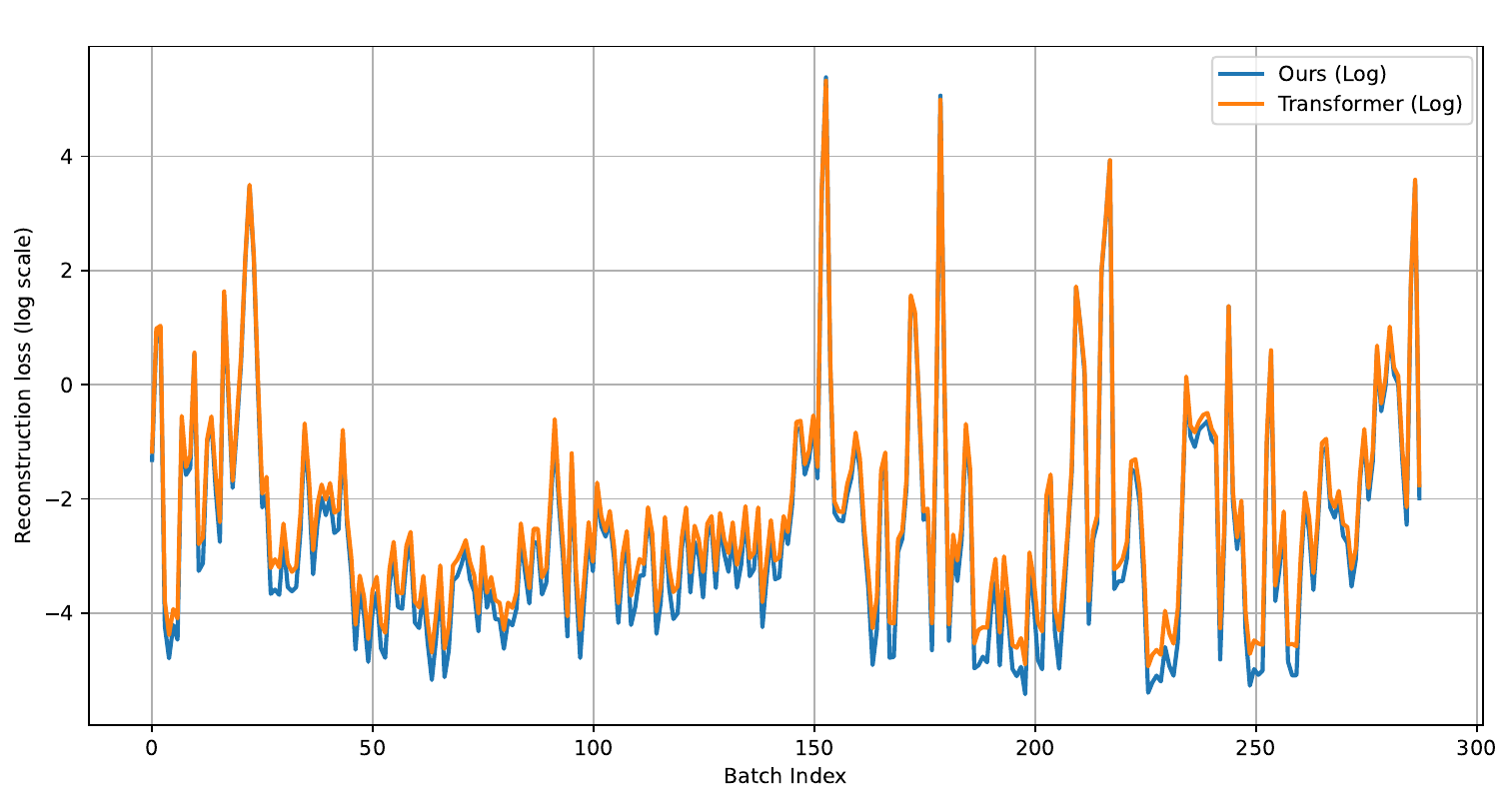}
        \caption{MSL Reconstruction Loss (Ours vs. Anomaly Transformer)}
        \label{fig:msl_loss}
    \end{subfigure}
    \hfill
    \begin{subfigure}{0.35\textwidth}
        \centering
        \includegraphics[width=\linewidth]{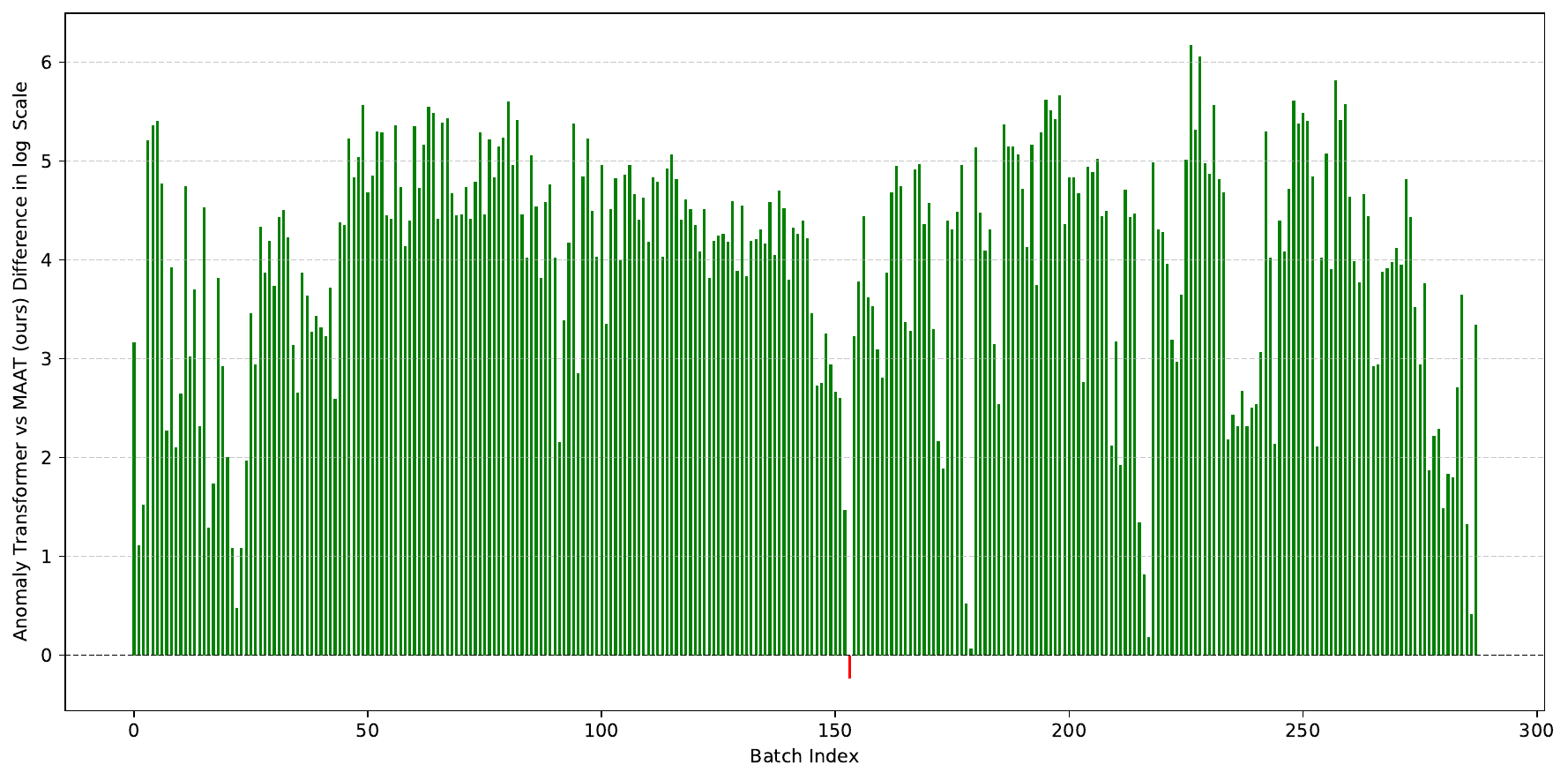}
        \caption{MSL Difference in Reconstruction Loss}
        \label{fig:msl_loss_diff}
    \end{subfigure}

    \vspace{0.5cm} 

    \begin{subfigure}{0.45\textwidth}
        \centering
        \includegraphics[width=\linewidth]{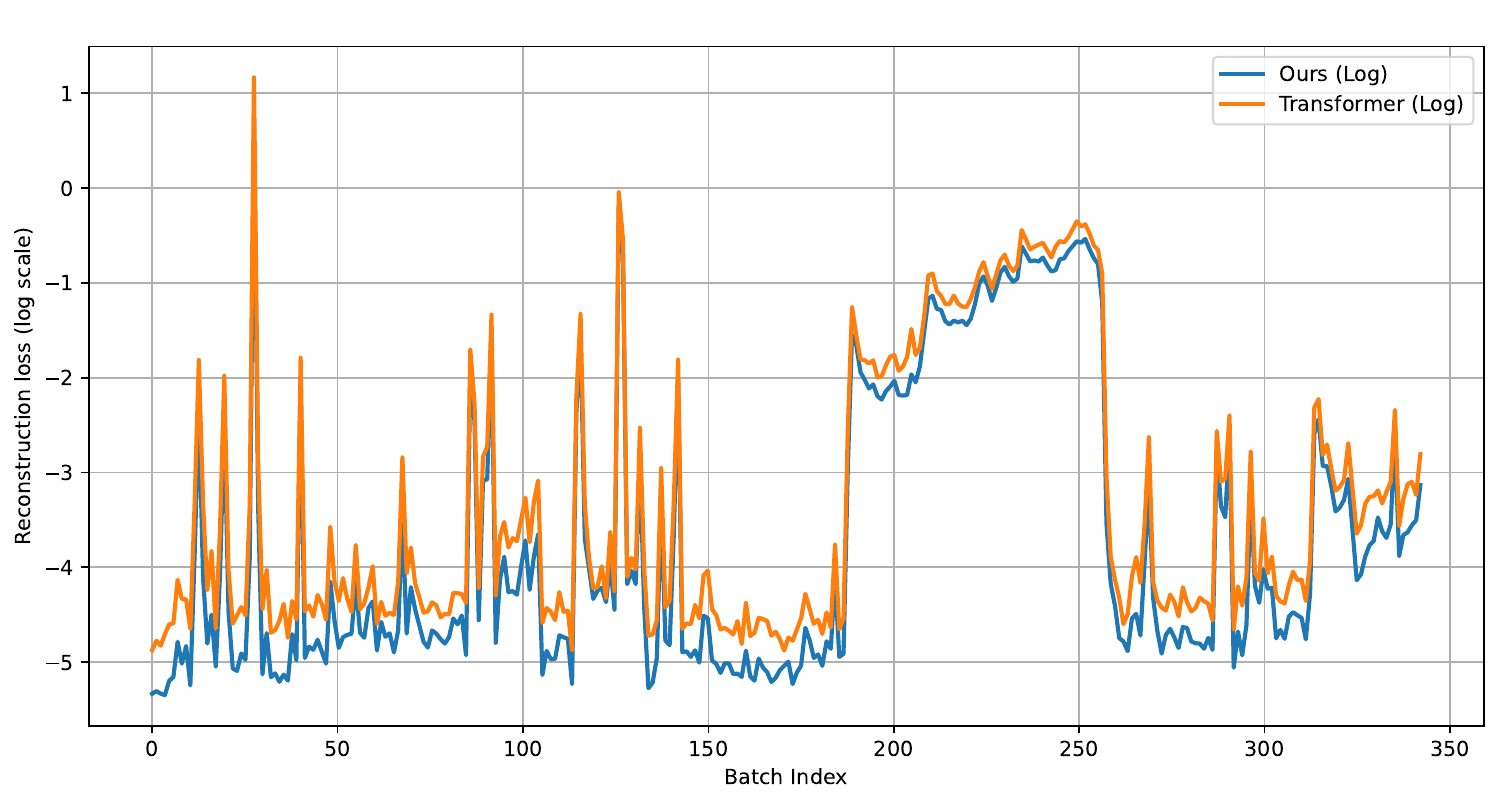}
        \caption{PSM Reconstruction Loss (Ours vs. Anomaly Transformer)}
        \label{fig:psm_loss}
    \end{subfigure}
    \hfill
    \begin{subfigure}{0.35\textwidth}
        \centering
        \includegraphics[width=\linewidth]{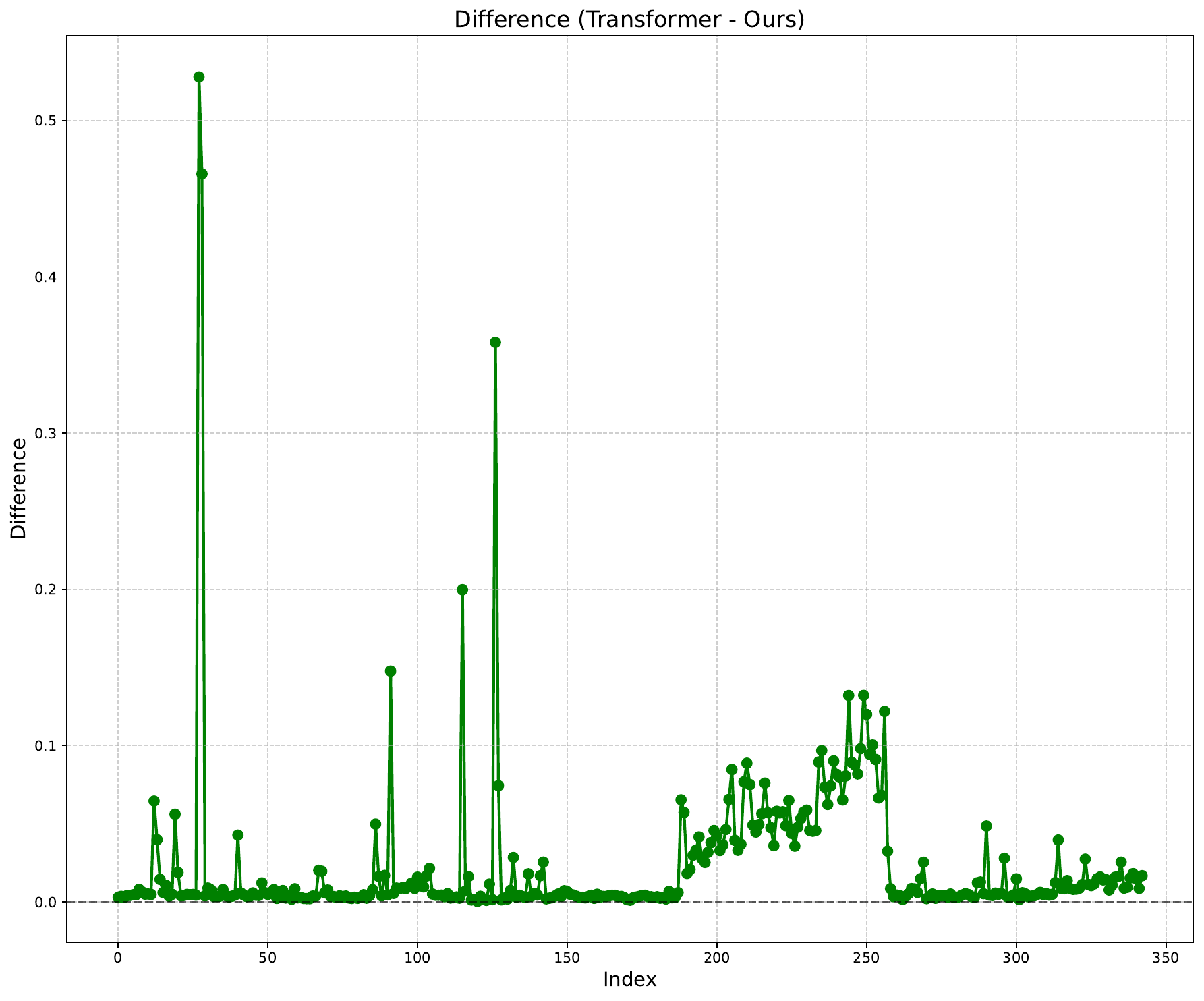}
        \caption{PSM Difference in Reconstruction Loss}
        \label{fig:psm_loss_diff}
    \end{subfigure}

    \vspace{0.5cm} 

    \begin{subfigure}{0.45\textwidth}
        \centering
        \includegraphics[width=\linewidth]{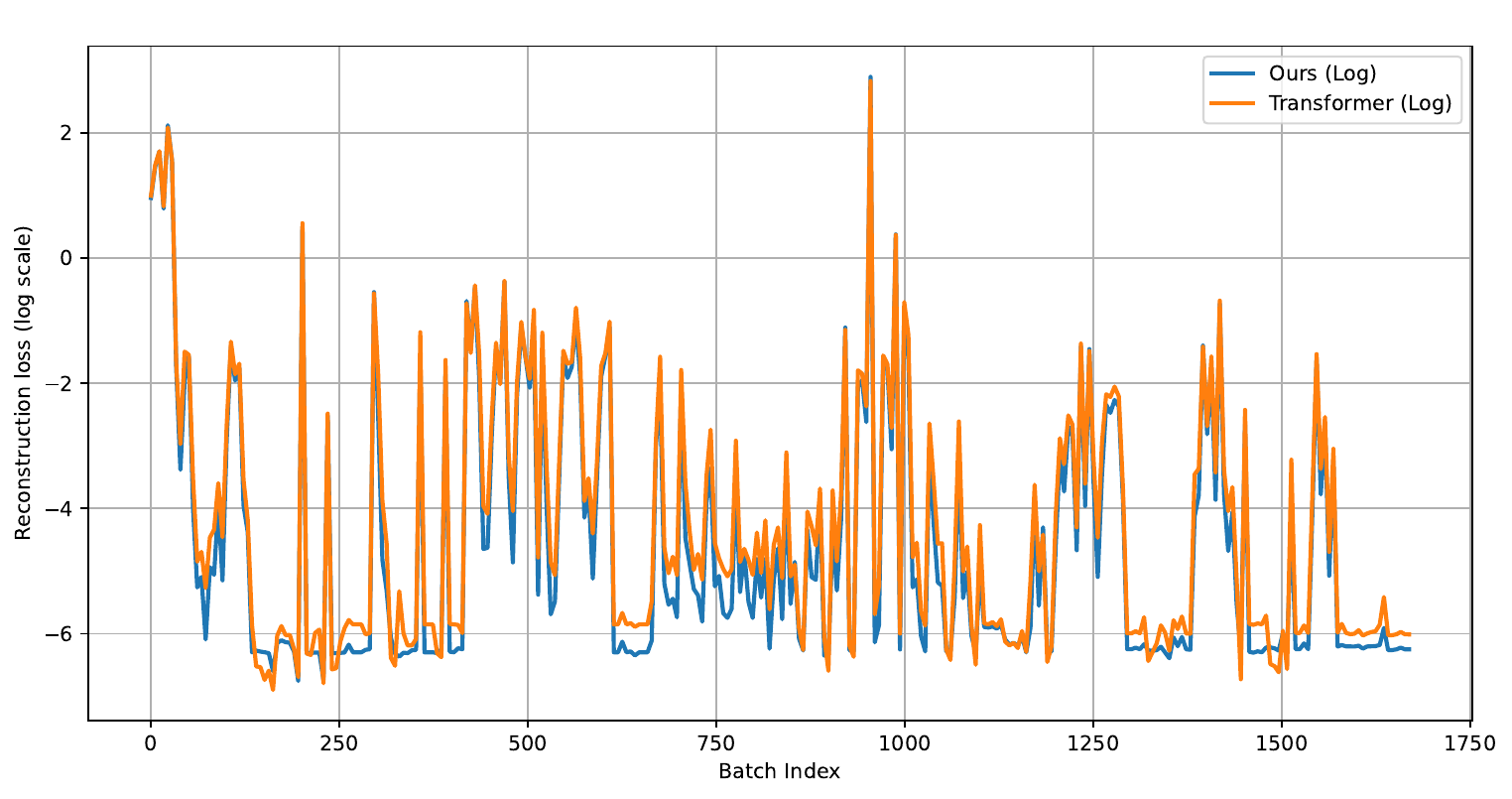}
        \caption{SMAP Reconstruction Loss (Ours vs. Anomaly Transformer)}
        \label{fig:smap_loss}
    \end{subfigure}
    \hfill
    \begin{subfigure}{0.35\textwidth}
        \centering
        \includegraphics[width=\linewidth]{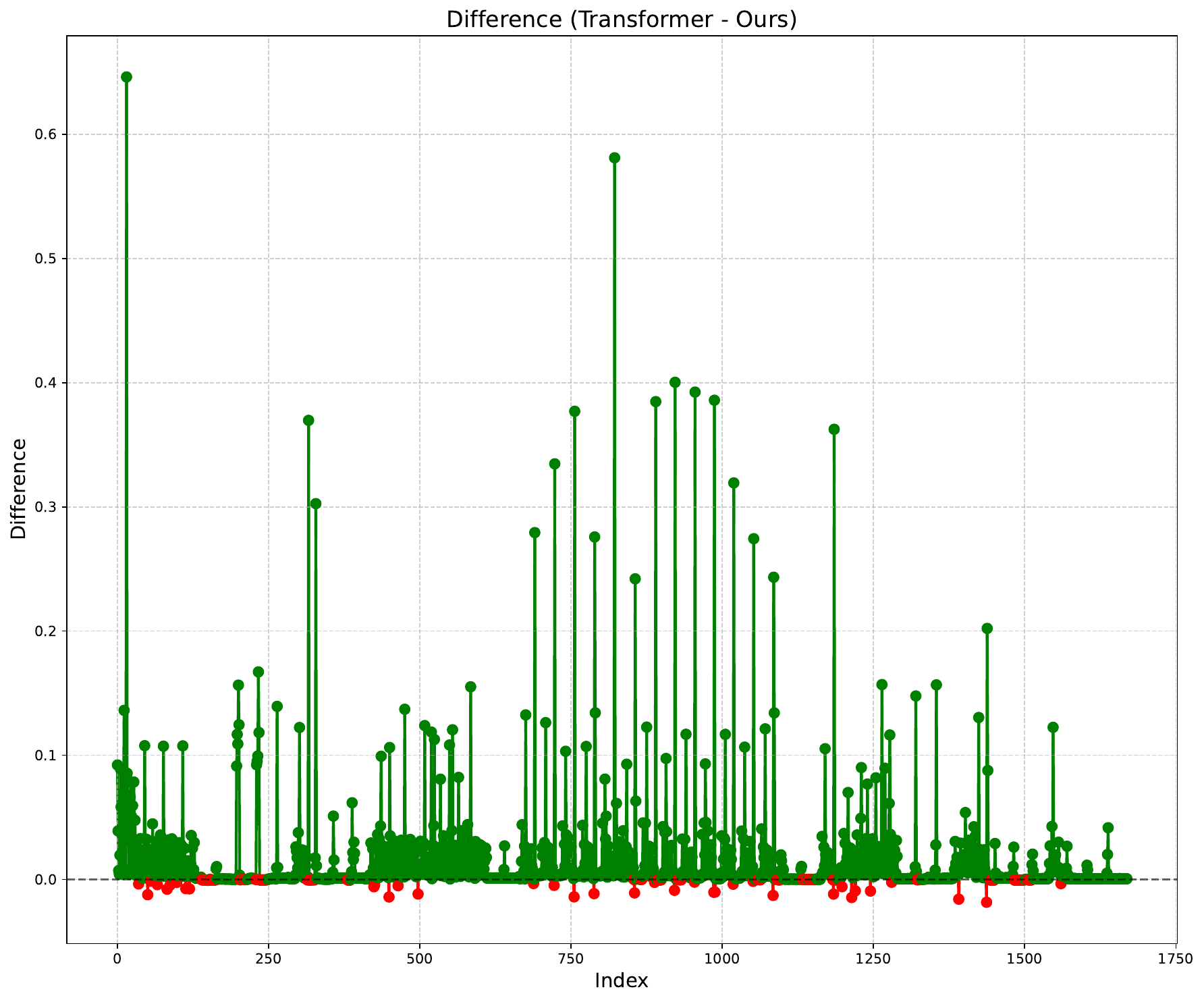}
        \caption{SMAP Difference in Reconstruction Loss}
        \label{fig:smap_loss_diff}
    \end{subfigure}

    \vspace{0.5cm} 

    \begin{subfigure}{0.45\textwidth}
        \centering
        \includegraphics[width=\linewidth]{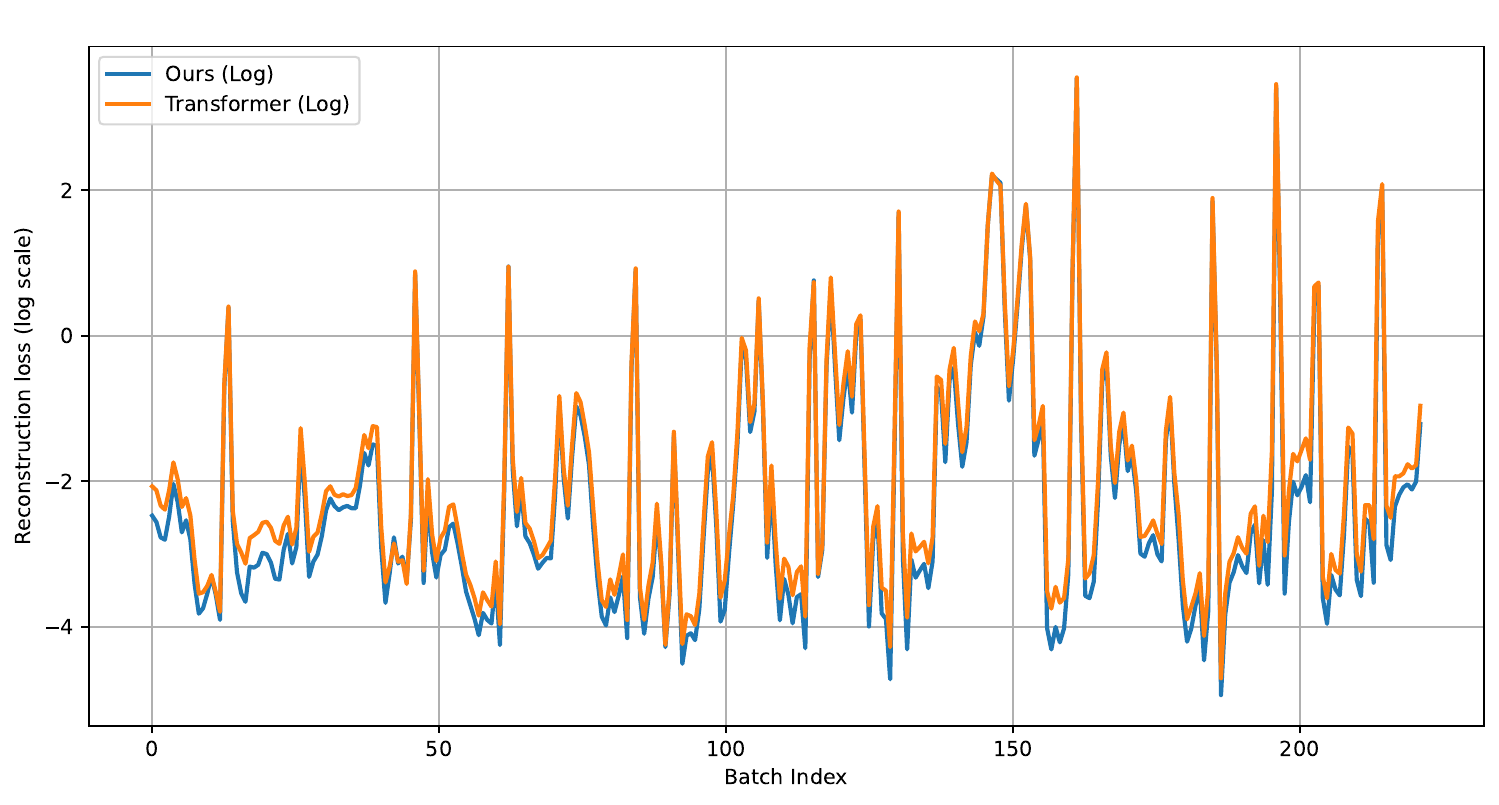}
        \caption{SMD Reconstruction Loss (Ours vs. Anomaly Transformer)}
        \label{fig:smd_32_loss}
    \end{subfigure}
    \hfill
    \begin{subfigure}{0.35\textwidth}
        \centering
        \includegraphics[width=\linewidth]{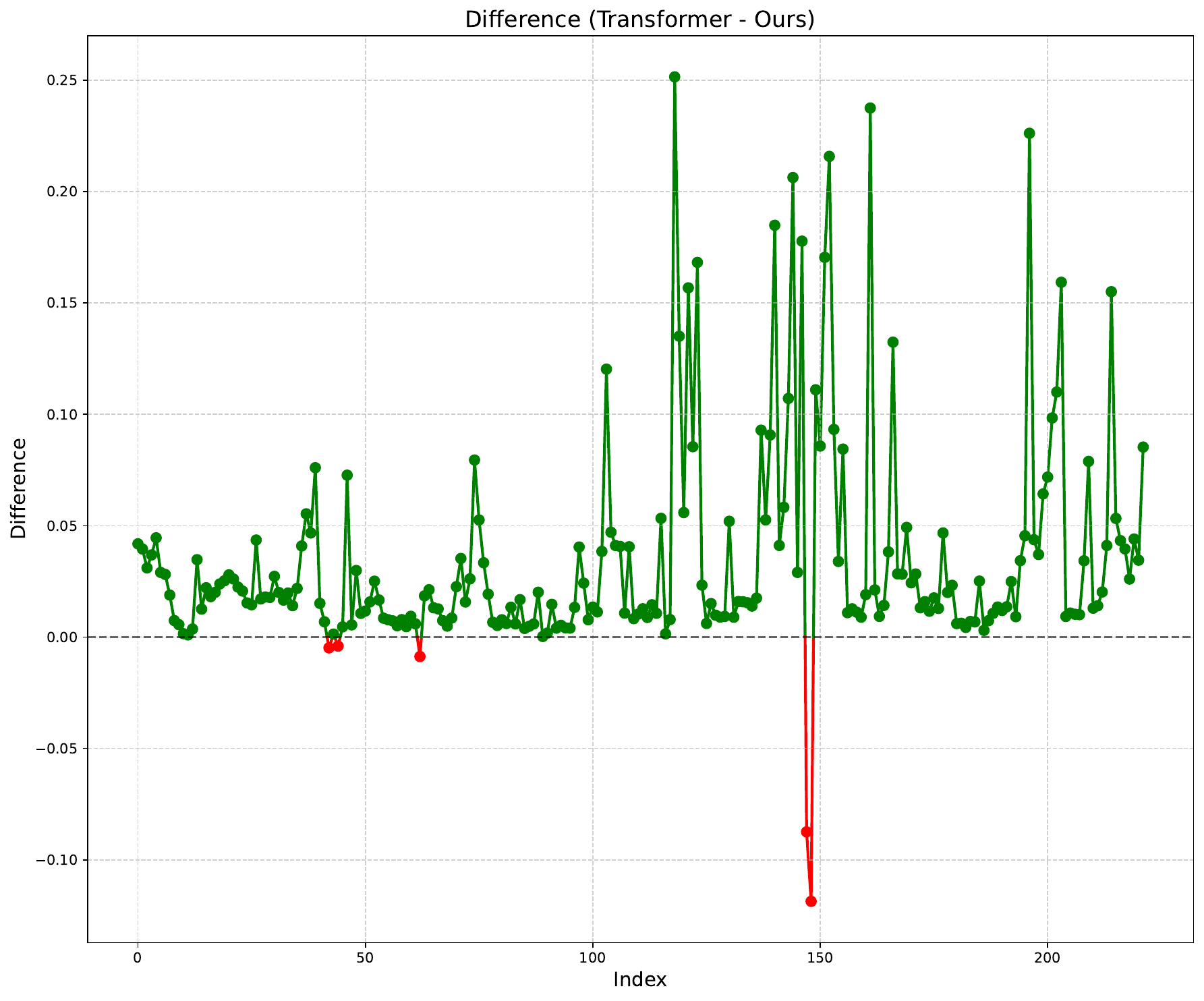}
        \caption{SMD Difference in Reconstruction Loss}
        \label{fig:smd_32_loss_diff}
    \end{subfigure}

    \caption{Comparison of reconstruction loss between our model and the Anomaly Transformer across different datasets. Left column: Reconstruction loss curves for both models. Right column: Difference in reconstruction loss.}
    \label{fig:all_graphs}
\end{figure}

\section{Implementation Details}

The MAAT model was implemented using \textbf{Python 3.8} and the \textbf{PyTorch} deep learning framework. PyTorch was chosen for its flexibility in defining custom architectures, efficient tensor operations, and strong support for GPU acceleration. The model was trained and evaluated on an \textbf{NVIDIA RTX Titan GPU with 24GB VRAM}, enabling efficient parallel computation for high-dimensional time-series data.  

To ensure stable and reproducible training, experiments were conducted in a controlled environment with fixed random seeds. The training process leveraged mixed-precision computation to optimize memory usage and speed, particularly beneficial for large-scale datasets such as SMD and SWaT. The implementation also utilized PyTorch’s \texttt{DataLoader} for efficient batch processing, automatic differentiation via \texttt{autograd}, and hardware acceleration with \texttt{torch.cuda}.  

The model was optimized using the \textbf{Adam optimizer} with weight decay regularization, and learning rate scheduling was applied to adapt to convergence dynamics. All experiments were run on a system equipped with \textbf{Intel Xeon CPUs} and \textbf{64GB RAM} to handle large dataset preprocessing and training workloads efficiently.

\section{MAAT algorithm} \label{alg:maat}
This appendix contains details about the MAAT block algorithm used in the study.
\begin{algorithm*}[]\caption{MAAT: Mamba Adaptive Anomaly Transformer with Sparse Attention}\label{alg:maat}\begin{algorithmic}[1]\Require Time series $x \in \mathbb{R}^{B \times L \times D}$\Require Parameters: $block\_size$, $d\_model$, $n\_heads$, $e\_layers$, $d\_state$, $d\_conv$\Ensure Reconstruction $x$, Anomaly scores: $series$, $prior$, $\sigma$
\State Initialize $series\_list \gets \emptyset$, $prior\_list \gets \emptyset$, $\sigma\_list \gets \emptyset$\State $x_{orig} \gets x$ \Comment{Preserve initial input}
\For{$i = 1$ to $e\_layers$} 
    \State \textbf{Sparse Attention Processing:}    \State $x, series, prior, \sigma \gets \text{sparse\_attn\_layer}(x, block\_size, attn\_mask)$    \Comment{Compute sparse block-wise attention over windows of size $block\_size$}
    \State \textbf{Mamba Skip Path:}    \State $x_{mamba} \gets \text{MambaBlock}(x)$ \Comment{State-space model capturing long-range dependencies}    \State $x_{skip} \gets x_{mamba} + x_{orig}$ \Comment{Residual connection}    \State $x_{skip} \gets \text{LayerNorm}(x_{skip})$
    \State \textbf{Adaptive Gating:}    \State $g \gets \sigma\big(\text{Linear}([x; x_{skip}])\big)$ \Comment{Gated feature fusion}    \State $x \gets g \odot x_{skip} + (1-g) \odot x$ \Comment{Learnable blend of skip and main path}
    \State \textbf{State Update:}    \State $x_{orig} \gets x$ \Comment{Update input for next layer}    \State Append $series$ to $series\_list$    \State Append $prior$ to $prior\_list$    \State Append $\sigma$ to $\sigma\_list$
\EndFor
\If{$\text{norm} \neq \text{None}$}    \State $x \gets \text{LayerNorm}(x)$\EndIf
\State \Return $x, series\_list, prior\_list, \sigma\_list$
\end{algorithmic}\end{algorithm*}




\end{document}